\documentclass[a4paper,fleqn]{cas-sc}
\usepackage{subfigure}
\usepackage{graphicx}
\usepackage{float}
\usepackage{threeparttable}
\usepackage{hyperref}
\usepackage{url}
\usepackage{diagbox}
\usepackage{ulem}
\usepackage{caption}
\usepackage{xcolor}
\usepackage{lineno}
\usepackage[authoryear]{natbib}

\def\tsc#1{\csdef{#1}{\textsc{\lowercase{#1}}\xspace}}
\tsc{WGM}
\tsc{QE}
\tsc{EP}
\tsc{PMS}
\tsc{BEC}
\tsc{DE}

\begin{document}
\begin{sloppypar}
\let\WriteBookmarks\relax
\def\floatpagepagefraction{1}
\def\textpagefraction{.001}
\shorttitle{Information Fusion}
\shortauthors{Ting Wang et~al.}


\title [mode = title]{Knowledge-data fusion oriented  traffic state estimation: A stochastic physics-informed deep learning approach} 

\author[1,2,3]{Ting Wang}[orcid=0000-0002-0460-0682]
\ead{2110763@tongji.edu.cn}

\author[1,2]{Ye Li}
\ead{JamesLI@tongji.edu.cn}

\author[4]{Rongjun Cheng}
\ead{chengrongjun@nbu.edu.cn}

\author[1,2]{Guojian Zou}
\ead{2010768@tongji.edu.cn}

\author[3]{Takao Dantsuji}
\ead{takao.dantsuji@monash.edu}

\author[3]{Dong Ngoduy}
\cormark[1]
\ead{Dong.Ngoduy@monash.edu}

\address[1]{The Key Laboratory of Road and Traffic Engineering, Ministry of Education, Tongji University, Shanghai 201804, China}
\address[2]{College of Transportation Engineering, Tongji University, Shanghai 201804, China}
\address[3]{Monash Institute of Transport Studies, Monash University, Melbourne VIC3800, Australia}
\address[4]{Faculty of Maritime and Transportation, Ningbo University, Ningbo 315211, China}

\begin{abstract}
Physics-informed deep learning (PIDL)-based models have recently garnered remarkable success in traffic state estimation (TSE). However, the prior knowledge used to guide regularization training in current mainstream architectures is based on deterministic physical models. The drawback is that a solely deterministic model fails to capture the universally observed traffic flow dynamic scattering effect, thereby yielding unreliable outcomes for traffic control. This study, for the first time, proposes stochastic physics-informed deep learning (SPIDL) for traffic state estimation. The idea behind such SPIDL is simple and is based on the fact that a stochastic fundamental diagram provides the entire range of possible speeds for any given density with associated probabilities. Specifically, we select percentile-based fundamental diagram and distribution-based fundamental diagram as stochastic physics knowledge, and design corresponding physics-uninformed neural networks for effective fusion, thereby realizing two specific SPIDL models, namely \text{$\alpha$}-SPIDL and \text{$\cal B$}-SPIDL. The main contribution of SPIDL lies in addressing the "overly centralized guidance" caused by the one-to-one speed-density relationship in deterministic models during neural network training, enabling the network to digest more reliable knowledge-based constraints.
Experiments on the real-world dataset indicate that proposed SPIDL models achieve accurate traffic state estimation in sparse data scenarios. More importantly, as expected, SPIDL models reproduce well the scattering effect of field observations, demonstrating the effectiveness of fusing stochastic physics model knowledge with deep learning frameworks.

\end{abstract}


\maketitle
\section{Introduction}
With the development and iteration of big data and cloud computing technologies, empowering traffic management and control with artificial intelligence (AI) has become the current research hotspot \citep{zou2023novel}. The premise of traffic analysis and control is to have a holographic perception and an in-depth grasp of the overall traffic states. However, due to factors such as sparse deployment of traffic sensors and adverse weather conditions, traffic data are often limited and noisy. Therefore, traffic state estimation (TSE) is of great significance as it can fill some gaps in spatiotemporal traffic information measurement and provide accurate and high-resolution global traffic estimation \citep{wang2022real}. 

Scholars have conducted and accumulated rich research on TSE. Among them, model-driven TSE methods are mainly based on prior knowledge of traffic dynamics described by physical models, which have high interpretability and portability \citep{ngoduy2011low}. However, the estimation results suffer from notable inaccuracies due to the limited expressive capacity of mathematical formulas. Besides, data-driven TSE models proficiently handle the nonlinear relationship between traffic states and factors like time and road topology, often outperforming model-driven approaches in accuracy \citep{wang2024spatiotemporal}. Although promising, data-driven methods lack stability and interpretability, which is their limitation as a black box. Is there a continued effort to simultaneously satisfy two seemingly contradictory pursuits (accuracy and interpretability)? The answer is physics-informed deep learning (PIDL). The application of PIDL in TSE has shown great potential \citep{shi2021physics, zhao2023observer, xue2024network}. PIDL employs small data for pattern discovery while respecting the physical laws by fusing a weakly imposed loss function.  

Almost all the existing PIDL-based TSE models were endowed with deterministic traffic flow models as physical prior knowledge, including fundamental diagrams (FDs) and macroscopic traffic flow models. Inside, FDs have been used as an essential input to macroscopic traffic flow models such as Lighthill-Witham-Richards (LWR) \citep{lighthill1955kinematic}, Aw-Rascle-Zhang (ARZ) \citep{aw2000resurrection} and cell transmission model \citep{daganzo1994cell}. However, despite the simplicity and practicality of the deterministic fundamental diagram \citep{greenshields1935study, underwood1960speed}, the widespread scattering effect observed in empirical data cannot be captured as reported in \citep{cheng2024analytical}. Simply put, a solely deterministic fundamental diagram lacks coverage of all observed samples. The deterministic speed-density model provides only one speed value for a given traffic density, often corresponding to the mean of the actual distribution. This results in the model being given unrealistic physical constraints, which we refer to as “overly centralized guidance”. Furthermore, we do not wish to construct a stochastic model by simply adding random noise or distribution assumptions to the parameters of the deterministic traffic flow models. As reported in \citep{ni2018modeling}, that involves certain stochastic elements in the model, rather than true stochastic models. The analytically deriving stochastic fundamental diagram from its sources is what we expect to fuse and extend into the PIDL, thereby explicitly capturing the estimation uncertainty.  

Inspired by the issues mentioned above, unlike previous PIDL models that adopt deterministic fundamental diagrams to implement first-order models or second-order models, we apply two stochastic fundamental diagrams to describe the relationship between traffic state variables (speed and density in this paper) and satisfy the conservation law. Following this direction, this paper proposes two stochastic physics-informed deep learning frameworks for traffic state estimation, named \text{$\alpha$}-SPIDL and \text{$\cal B$}-SPIDL. In contrast to the current PIDL models, the proposed SPIDL models enforce the underlying physics in the form of "soft constraints", instead of the existing "hard constraints" of deterministic fundamental diagrams, thereby expanding the ability to characterize uncertainties in traffic state estimation. The contributions of this paper are thus outlined as follows: 

\begin{itemize}
\item In \text{$\alpha$}-SPIDL-LWR and \text{$\alpha$}-SPIDL-ARZ, we apply the optimization model in \citep{qu2017stochastic} to calibrate a family of percentile-based speed-density curves, which were then combined with first-order LWR and second-order ARZ model as prior knowledge. Based on this, physical constraints are constructed to regularize the neural network training. 

\item In \text{$\cal B$}-SPIDL, we utilize a stochastic process modeling procedure from \citep{ni2018modeling} to obtain the speed distribution under varying density levels. Besides, we construct a variational autoencoder (VAE) network with a dual encoder-decoder architecture for generated and inferred speed and density. The network is ultimately optimized by fusing the data loss and physical distribution loss.

\item We conduct experiments in sparse detection data scenarios across different numbers of available loop detectors, and the results demonstrate that the proposed SPIDL models can achieve robust boundary estimation and accurately replicate the traffic flow evolution. 
\end{itemize} 

The remainder of the paper is organized as follows. Section \ref{s2} reviews and summarizes previous related work. Section \ref{s3} provides a detailed introduction of the specific models proposed within the stochastic physics-informed deep learning perspective. The experimental settings and results are presented in Section \ref{s4}. Section \ref{s5} concludes our work and prospects.

\section{Related work} \label{s2}
\begin{figure}[pos=htbp]
    \centering
\includegraphics[width=1\textwidth]{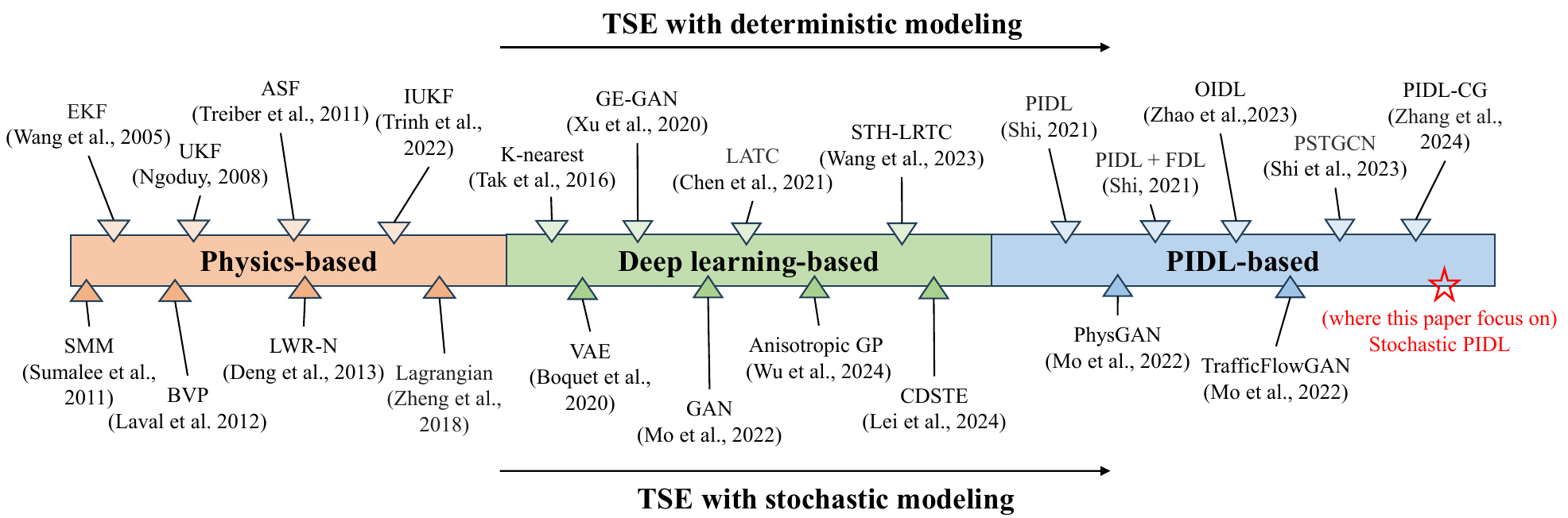}
    \caption{\centering{A schematic diagram of related literature}}
	\label{Relatedwork}
\end{figure}

This section succinctly reviews the related work on TSE with deterministic/stochastic modeling, with an overall summary depicted in Fig. \ref{Relatedwork}. Besides, some studies on stochastic modeling of traffic flow are introduced.

\subsection{TSE with deterministic modeling}
Classically, \citep{wang2005real} developed a general approach for TSE based on macroscopic traffic flow simulation and extended Kalman filters (EKF), which sparked a subsequent wave of TSE research around Kalman filters \citep{ngoduy2008applicable,trinh2022incremental,trinh2024stochastic}. Besides, methods such as particle filter \citep{wang2016efficient} and adaptive smoothing filter \citep{treiber2011reconstructing} also have scattered applications in TSE. Data-driven methods directly extract data correlations and patterns from historical data for TSE modeling. For example, k-nearest neighbors \citep{tak2016data} and tensor decomposition \citep{chen2021low, wang2023low} had been applied. Besides, some deep learning frameworks for TSE emerged, such as deep belief network \citep{tu2021estimating} and generative adversarial network (GAN) \citep{xu2020ge}. Recently, PIDL paradigm has opened up new ideas for TSE. \citep{huang2023incorporating} applied LWR model and nonlocal LWR model to the PIDL framework. In addition, the second-order traffic flow model had also been introduced into the PIDL \citep{shi2021physicsA}. In order to balance complexity and the trainability of physical encoding, \citep{shi2021physics} designed a fundamental graph learner into the PIDL to simultaneously handle traffic state estimation, model parameter recognition, and fundamental graph estimation. \citep{zhao2023observer} proposed the observer-informed deep learning (OIDL) model to estimate traffic states with boundary sensing data. Subsequently, spatiotemporal dependencies were also taken into account, and PIDL was extended to physics-informed spatiotemporal graph convolutional networks \citep{shi2023physics}. Lately, \citep{zhang2024physics} embedded fundamental graph parameters into the computational graph framework to obtain fundamental graph parameters. However, the above TSE methods with deterministic modeling have difficulty in handling the high degree of uncertainty in traffic state, which potentially arises from sources such as measurement error, dynamic behavior of various entities, etc.

\subsection{TSE with stochastic modeling}
Some early studies used stochastic models to capture estimation uncertainty \citep{jabari2013stochastic}, but mainly simply added random noise to the numerical format of deterministic traffic flow models. For example, \citep{sumalee2011stochastic} developed a LWR-based stochastic model. \citep{laval2012stochastic} expanded Newell’s principle to estimate traffic states with confidence intervals. \citep{ngoduy2021noise} developed a generic stochastic higher-order continuum model which follows a well-known Cox–Ingersoll–Ross (CIR) modeling process.  Besides, there are also some attempts for uncertainty quantification \citep{ kurzhanskiy2012guaranteed,deng2013traffic}. \citep{zheng2018traffic} interpreted the stochasticity in traffic flow models as one that describes uncertainty about the vehicle/driver attributes. Advancing into the era of deep learning, generative networks are the mainstream, including GAN \citep{mo2022quantifying, li2024self}, normalizing flow \citep{huang2023bridging}, and variational autoencoder (VAE) \citep{boquet2020variational}. These models do not rely on any noise assumptions, but require sufficient and balanced data, thus lacking the necessary scalability and flexibility. In addition, there are two interesting recent studies, anisotropic Gaussian processes \citep{wu2024traffic} and the conditional
diffusion framework with spatiotemporal estimator (CDSTE) \citep{lei2024conditional}. The former provides statistical uncertainty quantification for the imputed traffic
state with trajectory data, while the latter combines the frontier conditional diffusion framework to achieve reliable probabilistic estimation. Focusing on PIDL with stochastic modeling, to the best of our knowledge, there are mainly the following two groundbreaking studies, TrafficFlowGAN \citep{mo2022trafficflowgan} and PhysGAN-TSE \citep{mo2022quantifying}. TrafficFlowGAN uses a normalizing flow model as the generator to explicitly estimate the likelihood value while utilizing adversarial training with the discriminator to ensure sample quality. The core contribution of PhysGAN-TSE is to directly treat physics parameters as random variables and assume that the variables follow a conditional distribution. However, fusing the analytically deriving stochastic traffic flow models into the PIDL framework is still a gap that needs to be filled. 

\subsection{Stochastic modeling of traffic flow}
Realizing the limitation of the deterministic model, some studies have attempted to model stochastic fundamental diagram 
\citep{bai2021calibration, cheng2024analytical}. For example, \citep{ngoduy2011multiclass} considered scattering effects and used a multi-class first-order model with stochastic parameters settings. Afterwards, more studies began to analytically derive stochastic models, instead of adding random noise to the parameters of deterministic traffic flow models. Typically, \citep{qu2017stochastic} applied a total probability-based optimization model to calibrate the speed distributions with the given density. 
Subsequently, this work was also applied to model the behavior of heterogeneous and undisciplined traffic streams with UAV-based data \citep{ahmed2021fundamental}. As a continuation of \citep{qu2017stochastic}, a mean absolute error minimization-based holistic modeling framework was proposed \citep{wang2021model}, aiming to solve the previous problem that a family of percentile-based stochastic fundamental diagrams is inadequate to coordinate. From another perspective, \citep{ni2018modeling} modeled the phase diagrams as a stochastic process captured completely in a Maxwell–Boltzmann-like distribution, which is exactly the expected stochastic traffic flow model in the stochastic physics-informed deep learning framework of this paper.

\section{Methodology} \label{s3}
\subsection{Problem definition}
The traffic dynamics evolution of a road corridor with a length of $L$ is considered in this paper. The traffic state $s$ includes traffic speed $v\left( {x,t} \right)$, density $\rho \left( {x,t} \right)$ and flow $q\left( {x,t} \right) = \rho \left( {x,t} \right) \cdot v\left( {x,t} \right)$, with $t$ indicating the time of data collection and $x$ indicating the location of data collection. The complete spatiotemporal domain is ${\cal G} = \left\{ {\left( {x,t} \right)\left| {0 \le x \le L,0 \le t \le T} \right.} \right\}$. And the limited observed state data is defined as subset ${s_{\cal O}} \in {s_{\cal G}}$. And the problem of traffic state estimation is to learn a mapping function ${f_\theta }\left(  \cdot  \right)$ to infer the traffic state of the entire corridor ${s_{\cal G}}$ based on limited observations ${s_{\cal O}}$, that is ${s_{\cal G}} = {f_\theta }\left( {{s_{\cal O}}} \right)$. In this paper, the state is selected as density $\rho \left( {x,t} \right)$ and speed $v\left( {x,t} \right)$. Particularly, in the PIDL architecture, in addition to observed state space ${\cal O} \in {\cal G}$, it is necessary to define and construct a collocation state space ${\cal C} \in {\cal G}$ \citep{raissi2018deep}. The collocation point only provides a location-time vector $\left( {{x_{\cal C}},{t_{\cal C}}} \right) \in {\cal C}$ within the domain ${\cal G}$, observation and collocation points are used together as training data.

\subsection{Percentile-based SPIDL (\text{$\alpha$}-SPIDL)}
\subsubsection{Overall architecture}
\begin{figure}[pos=htbp]
    \centering
\includegraphics[width=1\textwidth]{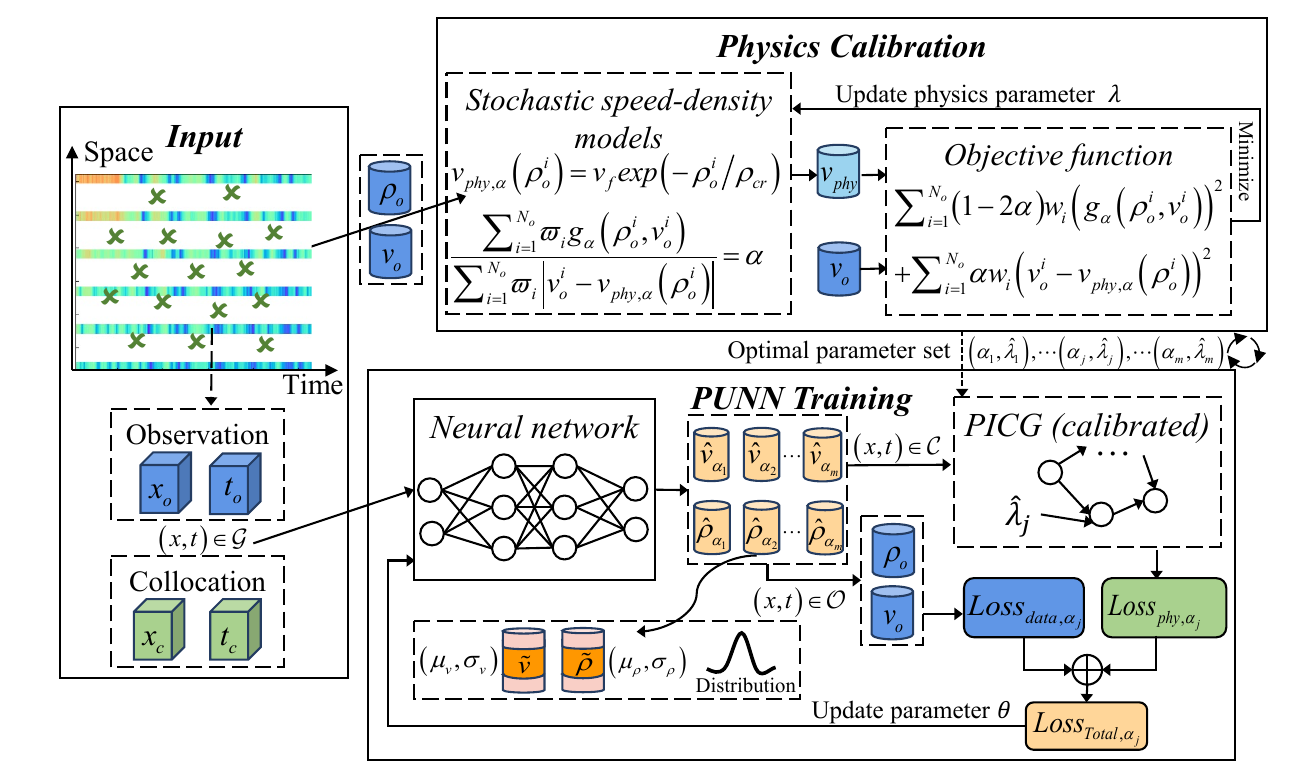}
    \caption{\centering{The architecture of percentile-based SPIDL (\text{$\alpha$}-SPIDL)}}
	\label{planA}
\end{figure}

The architecture of Percentile-based SPID (\text{$\alpha$}-SPIDL) is shown in Fig. \ref{planA}. Firstly, in the physics calibration layer, limited observation data is input into a percentile-based fundamental diagrams model \citep{qu2017stochastic}, and a family of optimal parameters corresponding to different percentages $\alpha  = \left( {{\alpha _1},{\alpha _2}, \cdots ,{\alpha _m}} \right)$ is obtained by minimizing the objective function. This process ensures the accuracy of model parameters and the effective coverage of observed traffic flow scattering effects by the model. Therefore, the speed distribution of any given density can be obtained. Then, the above set of optimal physics parameter $\left( {{\alpha _j},{{\hat \lambda }_j}} \right),j \in m$ calibrated offline above is passed to the neural network training layer. Correspondingly, the two feature variables of location and time $\left( {x,t} \right) \in {\cal G}$ are input into the neural network to output estimated density ${\hat \rho _{{\alpha _j}}}$ and speed ${\hat v_{{\alpha _j}}}$, and the difference between the estimated values and the observed values is used as the data loss $Los{s_{data,{\alpha _j}}}$. Meanwhile, the estimated speed and density of the collocation points are input into the physics-informed computational graph (PICG), and the error caused by the deviation of the estimated values from the traffic flow model is calculated as the physics loss $Los{s_{phy,{\alpha _j}}}$. Finally, the loss function $Los{s_{Total,{\alpha _j}}}$ used for training \text{$\alpha$}-SPIDL is composed of the data loss and physics loss. Following the above process, a family of physics parameters $({\alpha _1},{\hat \lambda _1}),({\alpha _2},{\hat \lambda _2}), \cdots ({\alpha _m},{\hat \lambda _m})$ is iterated to ultimately form the estimated speed distribution $\tilde v \sim ({\mu _v},{\sigma _v})$ and estimated density distribution $\tilde \rho  \sim ({\mu _\rho },{\sigma _\rho })$.

\subsubsection{Stochastic speed-density models}
Following the modeling process of the percentile-based fundamental diagrams \citep{qu2017stochastic, ahmed2021fundamental}, we first calibrate a family of percentile-based speed-density curves. The objective of the percentile-based model [M’] is to calibrate a $100\alpha th$ percentile-based speed-density curve. And $\alpha $ is defined as the ratio between weighted residual
of observations below the calibrated percentile-based curve and the total residual, expressed as follows:

\begin{equation}
\frac{{\sum\nolimits_{i = 1}^{{{N_{\cal O}}}} {{\varpi _i}{g_\alpha }\left( {\rho _o^i,v_o^i} \right)} }}{{\sum\nolimits_{i = 1}^{{{N_{\cal O}}}} {{\varpi _i}\left| {v_o^i - {v_{phy,\alpha }}\left( {\rho _o^i} \right)} \right|} }} = \alpha  
\end{equation}
where ${{N_{\cal O}}}$ is the total number of observation samples. Without loss of generality, we select Underwood model \citep{underwood1960speed} to estimate stochastic speed-density models in our framework. Thus the objective of the parameter calibration problem is defined as follows:
\begin{equation}
[M']minC\left( {{\rho _{cr}},{v_f}} \right) = \sum\nolimits_{i = 1}^{{{N_{\cal O}}}} {\left( {1 - 2\alpha } \right)} {w_i}{\left( {{g_\alpha }\left( {\rho _o^i,v_o^i} \right)} \right)^2} + \sum\nolimits_{i = 1}^{{{N_{\cal O}}}} \alpha  {w_i}{\left( {v_o^i - {v_{phy,\alpha }}\left( {\rho _o^i} \right)} \right)^2}
\end{equation}

Subject to:
\begin{equation}
\begin{array}{l}
{v_{phy,\alpha }}\left( {\rho _o^i} \right) = {v_f}{e^{\left( { - \frac{{\rho _o^i}}{{{\rho _{cr}}}}} \right)}}\quad i = 1,\;2,\;3,\; \cdots \;,{N_{\cal O}}\\
{g_\alpha }\left( {\rho _o^i,v_o^i} \right) = \left\{ \begin{array}{l}
{v_{phy,\alpha }}\left( {\rho _o^i} \right) - v_o^i,\quad if\;v_o^i - {v_{phy,\alpha }}\left( {\rho _o^i} \right) < 0\\
0,\quad \quad otherwise
\end{array} \right.\\
{\rho _{cr}} > 0\quad \& \quad {v_f} > 0
\end{array}
\end{equation}
where ${\rho _{cr}}$ and ${v_f}$ are critical density and free flow speed, ${w_i}$ is the weight that explains the sample selection bias, ${v_{phy,\alpha }}\left( {\rho _o^i} \right)$ is the estimated $100\alpha th$ percentile speed based on the selected stochastic speed-density function. Finally, we calibrated a family of optimal parameters $({\alpha _1},{\hat \lambda _1}), \cdots ,({\alpha _j},{\hat \lambda _j}), \cdots ,({\alpha _m},{\hat \lambda _m})$ with different $\alpha$, including optimal critical density and free flow speed, and fifteen $\alpha$ (see Table \ref{Calibration} for details) is selected in \text{$\alpha$}-SPIDL, that is $m = 15$.

\subsubsection{PUNN for deep learning module}
The essence is to achieve end-to-end learning from the spatiotemporal coordination $X = \left( {x,t} \right)$ to  the traffic states labels $s = \left[ {\rho \left( {x,t} \right),v\left( {x,t} \right)} \right]$. Thus, a neural network (NN) is designed with position $x$ and time $t$ as inputs, and the outputs are estimated traffic density ${\hat \rho}\left( {x,t;\theta } \right)$ and speed ${\hat v}\left( {x,t;\theta } \right)$, which can be expressed as:
\begin{equation}
\hat s = {f_\theta }\left( {x,t} \right) = {W_o}\sigma \left( {{W_h}X + {b_h}} \right) + {b_o}
\end{equation}
where $\theta $ represents trainable parameters in the NN, ${W_o}$ are ${{W_h}}$ the weight matrices of the output layer and the hidden layer, respectively, $\sigma $ is Activation 
function, ${{b_h}}$ and ${{b_o}}$ are the deviation matrices of the hidden layer and output layer. 

And we denote ${\hat \rho}\left( {x,t;\theta } \right)$ and speed ${\hat v}\left( {x,t;\theta } \right)$ as ${\hat \rho _{\alpha_j}}$ and ${\hat v_{\alpha_j}}$ when the corresponding optimal parameter input is $\left( {{\alpha _j},{{\hat \lambda }_j}} \right)$. Furthermore, the NN loss is calculated at the observation point $\left( {x,t} \right) \in {\cal O}$ to measure the difference between the observed state and the estimated state. The data loss is defined as the mean square error (MSE), that is:
\begin{equation}
Los{s_{data,{\alpha _j}}}\left( \theta  \right) = {\beta _\rho }\frac{1}{{{N_{\cal O}}}}\sum\limits_{i = 1}^{{N_{\cal O}}} {{{\left| {{{\hat \rho }_{i,{\alpha _j}}} - {\rho _o}} \right|}^2}}  + {\beta _v}\frac{1}{{{N_{\cal O}}}}\sum\limits_{i = 1}^{{N_{\cal O}}} {{{\left| {{{\hat v}_{i,{\alpha _j}}} - {v_o}} \right|}^2}} \quad (j = 1,2,3, \cdots m)
\end{equation}
where ${{N_{\cal O}}}$ is the sample size of observation points,  ${\beta _\rho }$ and ${\beta _v}$ are weight coefficients used to balance the contributions of density loss and speed loss, respectively.

\subsubsection{Knowledge used for regularization training}
Here, the physical knowledge is embedded in the NN by substituting the estimated results into the traffic flow model to check if they match the traffic flow model as prior knowledge. Note that the previously obtained family of speed-density curves is the key for \text{$\alpha$}-SPIDL to achieve traffic state uncertain estimation. This gives us different physical parameters, allowing us to construct different physical constraints, iterate to obtain different estimation results, and generate the final estimation distribution. 

Specifically, we applied two representative traffic flow models, the first-order LWR model \citep{lighthill1955kinematic} and the second-order ARZ model \citep{aw2000resurrection}, in our \text{$\alpha$}-SPIDL framework for the illustration purposes. The LWR model describes the evolution of traffic density ${\rho \left( {x,t} \right)}$ and speed ${v\left( {x,t} \right)}$ as:
\begin{equation}
{\partial _t}\rho  + {\partial _x}\left( {\rho v} \right) = 0
\end{equation}

ARZ model further increases the description of speed changes, taking into account non-equilibrium traffic dynamics, as shown below:
\begin{equation}
\begin{array}{l}
{\partial _t}\rho  + {\partial _x}\left( {\rho v} \right) = 0\\
{\partial _t}\left( {v + p\left( \rho  \right)} \right) + v{\partial _x}\left( {v + p\left( \rho  \right)} \right) = \frac{{\left( {{V_e}\left( \rho  \right) - v} \right)}}{\tau }\\
p\left( \rho  \right) = {V_e}\left( 0 \right) - {V_e}\left( \rho  \right)
\end{array}
\end{equation}
where ${p\left( \rho  \right)}$ is  traffic pressure, ${{V_e}\left( \rho  \right)}$ is equilibrium speed and ${V_e}\left( 0 \right)$ is maximum speed (free flow speed ${{v_f}}$). 

Based on Eq.(3) and Eq.(6), the physics residual for \text{$\alpha$}-SPIDL-LWR can be defined as:
\begin{equation}
Los{s_{phyL,{\alpha _j}}} = {\gamma _1}\frac{1}{{{N_{\cal C}}}}\sum\limits_{i = 1}^{{N_{\cal C}}} {{{\left| {{{\hat v}_{i,{\alpha _j}}} - {v_{f,{\alpha _j}}}{e^{\left( { - \frac{{{{\hat \rho }_{i,{\alpha _j}}}}}{{{\rho _{cr,{\alpha _j}}}}}} \right)}}} \right|}^2}}  + {\gamma _2}\frac{1}{{{N_{\cal C}}}}\sum\limits_{i = 1}^{{N_{\cal C}}} {{{\left| {{\partial _t}{{\hat \rho }_{i,{\alpha _j}}} + {\partial _x}\left( {{{\hat \rho }_{i,{\alpha _j}}}{{\hat v}_{i,{\alpha _j}}}} \right)} \right|}^2}\quad } (j = 1,2,3, \cdots m)\label{phylwr}
\end{equation}
where $\gamma _1$ and $\gamma _2$ are weight coefficients.

Similarly, based on Eq.(3) and Eq.(7), the physics residual for  \text{$\alpha$}-SPIDL-ARZ is:
\begin{equation}
\begin{array}{l}
Los{s_{phyA,{\alpha _j}}} = {\eta _1}\frac{1}{{{N_{\cal C}}}}\sum\limits_{i = 1}^{{N_{\cal C}}} {{{\left| {{{\hat v}_{i,{\alpha _j}}} - {v_{f,{\alpha _j}}}{e^{\left( { - \frac{{{{\hat \rho }_{i,{\alpha _j}}}}}{{{\rho _{cr,{\alpha _j}}}}}} \right)}}} \right|}^2}}  + {\eta _2}\frac{1}{{{N_{\cal C}}}}\sum\limits_{i = 1}^{{N_{\cal C}}} {{{\left| {{\partial _t}{{\hat \rho }_{i,{\alpha _j}}} + {\partial _x}\left( {{{\hat \rho }_{i,{\alpha _j}}}{{\hat v}_{i,{\alpha _j}}}} \right)} \right|}^2}} \\
\quad \quad \quad \quad \quad  + {\eta _3}\sum\limits_{i = 1}^{{N_{\cal C}}} {\frac{1}{{{N_{\cal C}}}}{{\left| {{\partial _t}\left( {{{\hat v}_{i,{\alpha _j}}} + p\left( {{{\hat \rho }_{i,{\alpha _j}}}} \right)} \right) + {{\hat v}_i}{\partial _x}\left( {{{\hat v}_{i,{\alpha _j}}} + p\left( {{{\hat \rho }_{i,{\alpha _j}}}} \right)} \right) - \frac{{\left( {{V_e}\left( {{{\hat \rho }_{i,{\alpha _j}}}} \right) - {{\hat v}_{i,{\alpha _j}}}} \right)}}{\tau }} \right|}^2}} \quad (j = 1,2,3, \cdots m)
\end{array}\label{phyarz}
\end{equation}
where $\eta _1$, $\eta _2$ and $\eta _3$ are weight coefficients.

\subsubsection{Combined loss function}
The NN within the \text{$\alpha$}-SPIDL is optimized from both data-driven and model-based perspectives. Therefore, the parameter of NN is updated and iterated based on the following combined loss function:
\begin{equation}
\begin{array}{l}
Los{s_{Total,{\alpha _j}}}\left( {{\theta _L}} \right) = Los{s_{data,{\alpha _j}}} + Los{s_{phyL,{\alpha _j}}}\;\;\;({\rm{for}}\;\;{\rm{\alpha  - SPIDL - LWR}})\\
Los{s_{Total,{\alpha _j}}}\left( {{\theta _A}} \right) = Los{s_{data,{\alpha _j}}} + Los{s_{phyA,{\alpha _j}}}\;\;\;({\rm{for}}\;\;{\rm{\alpha  - SPIDL - ARZ}})
\end{array}
\end{equation}
where ${{\theta _L}}$ and ${{\theta _A}}$ are the parameters to be learned for \text{$\alpha$}-SPIDL-LWR model and \text{$\alpha$}-SPIDL-ARZ model, respectively.

\subsection{Distribution-based SPIDL (\text{$\cal B$}-SPIDL)}
\subsubsection{Overall architecture}
\begin{figure}[pos=htbp]
    \centering
\includegraphics[width=1\textwidth]{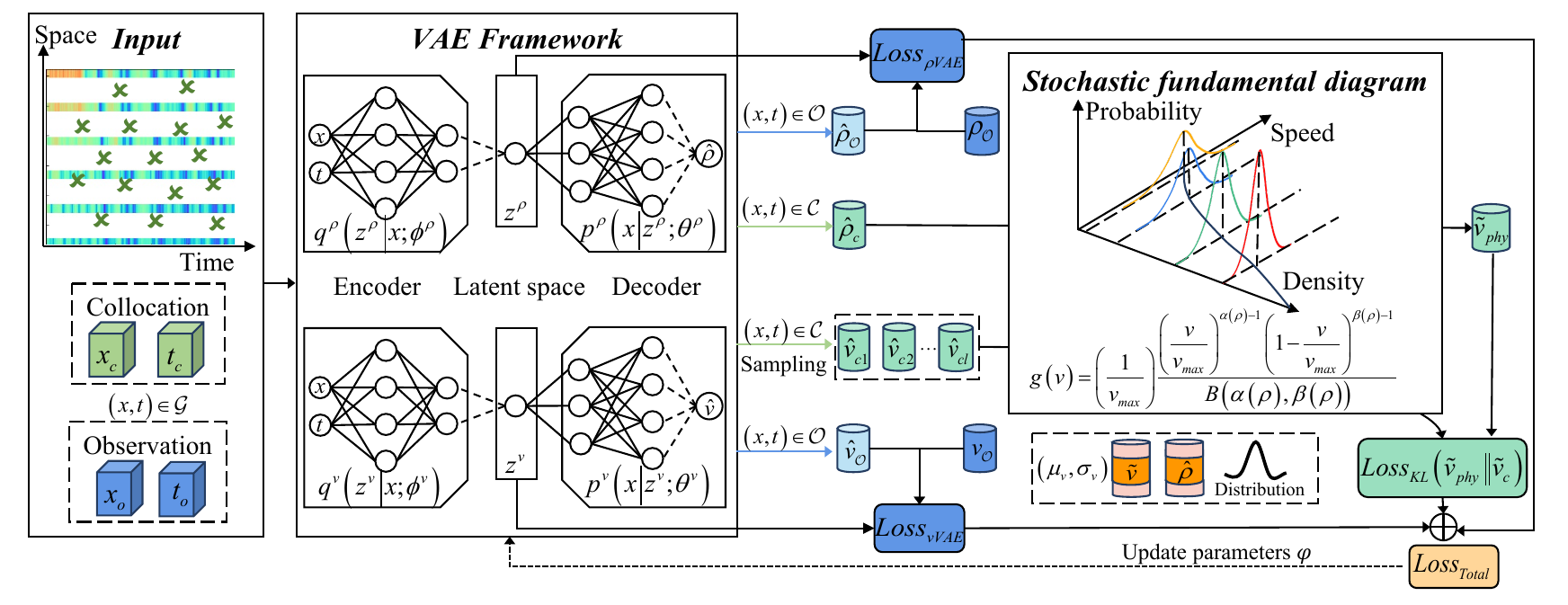}
    \caption{\centering{The architecture of distribution-based SPIDL (\text{$\cal B$}-SPIDL) }}
	\label{planB}
\end{figure}

From another perspective, Beta distribution-based SPIDL (\text{$\cal B$}-SPIDL)  directly generates the distribution of estimation based on VAE and then approximates it to the physical stochastic process. The architecture of \text{$\cal B$}-SPIDL is shown in Fig. \ref{planB}. Firstly, the spatiotemporal information $X = \left( {x,t} \right)$ is input into a VAE network with two pairs of encoder-decoder, and the estimated state values are output. Note that the VAE network used for density does not sample and outputs density estimations $\hat \rho $, while the VAE network used for speed requires sampling and outputs the distribution of speed estimations ${\tilde v_{\cal C}} = \left( {{{\hat v}_{{\cal C}1}},{{\hat v}_{{\cal C}2}}, \cdots ,{{\hat v}_{{\cal C}l}}} \right)$. Furthermore, the estimated states corresponding to the collocation points $\left( {x,t} \right) \in {\cal C}$ are fed into a Beta distribution-based stochastic fundamental \citep{ni2018modeling}, and the physical distribution Kullback-Leibler (KL) divergence ${D_{KL}}\left( {{{\tilde v}_{\cal C}}\left\| {{{\tilde v}_{phy}}} \right.} \right)$ is calculated to represent the deviation of speed distributions under varying density levels. Meanwhile, the deviation between the estimated state ${\hat \rho _{\cal O}},{\hat v_{\cal O}}$ corresponding to the observation point $\left( {x,t} \right) \in {\cal O}$ and the ground truth ${\rho _{\cal O}},{v_{\cal O}}$ is considered as the data loss of the VAE neural network $Los{s_{\rho net}},Los{s_{vnet}}$. Finally, the combination of stochastic process physical loss $Los{s_{phy}}$ and VAE neural network loss is used as the total loss $Los{s_{Total}}$ to train the network and achieve parameter updates. 

\subsubsection{The fundamental diagram as a stochastic process} \label{sp}
The modeling steps of the stochastic fundamental diagram follow the procedure presented in \citep{ni2018modeling}. Firstly, We discretize the entire continuous range into intervals and divide all observations into each interval, the number of intervals is set to $I$.  For interval $i \in I$, there are three variables: sample number ${n_i}$, mean ${\bar v_i}$ and variance ${w_i}$. Secondly, a s-shaped three-parameter traffic stream model \citep{cheng2021s} is used to establish an equilibrium speed-density relationship and fit the speed mean:  
\begin{equation}
v = \frac{{{v_f}}}{{{{\left[ {1 + {{\left( {{\rho  \mathord{\left/
 {\vphantom {\rho  {{\rho _{cr}}}}} \right.
 \kern-\nulldelimiterspace} {{\rho _{cr}}}}} \right)}^m}} \right]}^{{2 \mathord{\left/
 {\vphantom {2 m}} \right.
 \kern-\nulldelimiterspace} m}}}}}
\end{equation} 
where $v_f$ is the free flow speed, $\rho _{cr}$ is the critical density corresponding to the capacity,  $m$ is maximum flow inertia coefficient. Then, considering the evolution characteristics of traffic flow, that is, the speed variance is apparently small when density is low, and the speed variance peaks around optimal density, a log-normal density function with double-tailed peaks is selected to fit the variance of speed. Furthermore, we model the speed distribution along varying density levels. Following the analysis and comprehensive consideration of previous research \citep{ni2018modeling}, Beta distribution stands out and is selected as the distribution type. The probability density function (p.d.f.) of the Beta distribution is controlled by two positive parameters, $\alpha$ and $\beta$, which determine the shape of the distribution, as shown below:

\begin{equation}
f\left( x \right) = \frac{{{x^{\alpha  - 1}}{{\left( {1 - x} \right)}^{\beta  - 1}}}}{{B\left( {\alpha ,\beta } \right)}}\;\;\;with\;\;B\left( {\alpha ,\beta } \right) = \frac{{\Gamma \left( \alpha  \right)\Gamma \left( \beta  \right)}}{{\Gamma \left( {\alpha  + \beta } \right)}}
\end{equation} 
where $B\left( {\alpha ,\beta } \right)$ is the Beta function used for normalization to ensure a total probability of 1, and $\Gamma  $ represents the gamma function. The expected value and variance of the Beta distribution are given by the following formula:

\begin{equation}
\begin{array}{l}
E\left( X \right) = \frac{\alpha }{{\alpha  + \beta }}\\
Var\left( X \right) = \frac{{\alpha \beta }}{{{{\left( {\alpha  + \beta } \right)}^2}\left( {\alpha  + \beta  + 1} \right)}}
\end{array}
\end{equation}

Considering that the Beta distribution applies within $x \in \left[ {0,1} \right]$, the standard probability density function is transformed to be applicable to our scenario. The detailed transformation process can refer to \citep{ni2018modeling} and will not be elaborated here. The new p.d.f. associated with a speed range of $v \in \left[ {0,{v_{max}}} \right]$ as follows:

\begin{equation}
\begin{array}{l}
g\left( v \right) = \left( {\frac{1}{{{v_{max}}}}} \right)\frac{{{{\left( {\frac{v}{{{v_{max}}}}} \right)}^{\alpha \left( \rho  \right) - 1}}{{\left( {1 - \frac{v}{{{v_{max}}}}} \right)}^{\beta \left( \rho  \right) - 1}}}}{{B\left( {\alpha \left( \rho  \right),\beta \left( \rho  \right)} \right)}}\\
\alpha \left( \rho  \right) = \frac{{\hat \mu \left( \rho  \right)}}{{\hat \omega \left( \rho  \right)}}\left( {\hat \mu \left( \rho  \right) - {{\hat \mu }^2}\left( \rho  \right) - \hat \omega \left( \rho  \right)} \right)\\
\beta \left( \rho  \right) = \frac{{1 - \hat \mu \left( \rho  \right)}}{{\hat \omega \left( \rho  \right)}}\left( {\hat \mu \left( \rho  \right) - {{\hat \mu }^2}\left( \rho  \right) - \hat \omega \left( \rho  \right)} \right)
\end{array}
\label{pdf}
\end{equation}
where ${\hat \mu \left( \rho  \right)}$ and ${\hat \omega \left( \rho  \right)}$ are the mean and variance fitted from empirical observations, respectively. 

So far, the fundamental diagram modeling as a stochastic process over $\rho  \in \left[ {0,{\rho _{cr}}} \right]$ is complete. 

\subsubsection{VAE for distributed estimation}
The VAE framework in proposed includes two pairs of encoder-decoder for speed estimation and density estimation, respectively. Two neural network encoders ${q^\rho }\left( {{z^\rho }\left| {x;{\phi ^\rho }} \right.} \right),{q^v}\left( {{z^v}\left| {x;{\phi ^v}} \right.} \right)$ are recognition models, and the decoders ${p^\rho }\left( {x\left| {{z^\rho };{\theta ^\rho }} \right.} \right),{p^v}\left( {x\left| {{z^v};{\theta ^v}} \right.} \right)$ are generation models. 

\textbf{Encoders.} The density and speed encoders are implemented using MLP with weights and biases ${\phi ^\rho },{\phi ^v}$:
\begin{equation}
\begin{array}{l}
h_1^{\rho ,enc} = \sigma \left( {W_1^{\rho ,enc}X + b_1^{\rho ,enc}} \right),\;\;h_1^{v,enc} = \sigma \left( {W_1^{v,enc}X + b_1^{v,enc}} \right)\\
h_l^{\rho ,enc} = \sigma \left( {W_l^\rho h_{l - 1}^{\rho ,enc} + b_l^\rho } \right),\;\;h_l^{v,enc} = \sigma \left( {W_l^vh_{l - 1}^{v,enc} + b_l^v} \right),l = 2,..,L
\end{array}
\end{equation}
where $\sigma $ is the activation function, $L$ is the layer number of encoders. The outputs are the latent representation of the data, that is, the mean $\mu _z^\rho ,\mu _z^v$ and standard deviation $\sigma _z^\rho ,\sigma _z^v$: 
\begin{equation}
\begin{array}{l}
\mu _z^\rho  = W_\mu ^\rho h_L^{\rho ,enc} + b_\mu ^\rho ,\;\;\mu _z^v = W_\mu ^vh_L^{v,enc} + b_\mu ^v\\
\sigma _z^\rho  = \sigma \left( {W_\sigma ^\rho h_L^{\rho ,enc} + b_\sigma ^\rho } \right) + b_{low}^\rho ,\;\;\sigma _z^v = \sigma \left( {W_\sigma ^vh_L^{v,enc} + b_\sigma ^v} \right) + b_{low}^v
\end{array}
\end{equation}
where $b_{low}^\rho $ and $b_{low}^v$ are lower bounds added to help stabilize training. 

Thus, the stochastic representation ${z^\rho },{z^v}$ within latent space of density and speed networks are obtained by:
\begin{equation}
{z^\rho } = \mu _z^\rho  + {\delta ^\rho } \odot \sigma _z^\rho ,\;\;{z^v} = \mu _z^v + {\delta ^v} \odot \sigma _z^v
\end{equation}
where $\delta$ means noisy.

\textbf{Decoders.} Correspondingly, two decoders are implemented using MLP with weights and biases ${\theta ^\rho },{\theta ^v}$.   
\begin{equation}
\begin{array}{l}
h_1^{\rho ,dec} = \sigma \left( {W_1^{\rho ,dec}z + b_1^{\rho ,dec}} \right),h_1^{v,dec} = \sigma \left( {W_1^{v,dec}z + b_1^{v,dec}} \right)\\
h_m^{\rho ,dec} = \sigma \left( {W_{m - 1}^{\rho ,dec}h_{m - 1}^{\rho ,dec} + b_m^{\rho ,dec}} \right),h_m^{v,dec} = \sigma \left( {W_{m - 1}^{v,dec}h_{m - 1}^{v,dec} + b_m^{v,dec}} \right),m = 2,...,M\\
\hat \rho  = W_M^{\rho ,dec}h_M^{\rho ,dec} + b_M^\rho ,\hat v = W_M^{v,dec}h_M^{v,dec} + b_M^v
\end{array}
\end{equation}

Finally, the output of VAE network is $\hat \rho ,\hat v$, i.e., the data-driven density and speed estimation.

As for the loss function of VAE, it consists of distribution loss and prediction loss:
\begin{equation}
\begin{array}{l}
los{s_{\rho VAE}} = {D_{KL}}\left( {{q^\rho }\left( {{z^\rho }\left| X \right.} \right)\left\| {{p^\rho }\left( {{z^\rho }} \right)} \right.} \right) + MSE\left( {\hat \rho ,{\rho _{\cal O}}} \right)\\
los{s_{vVAE}} = {D_{KL}}\left( {{q^v}\left( {{z^v}\left| X \right.} \right)\left\| {{p^v}\left( {{z^v}} \right)} \right.} \right) + MSE\left( {\hat v,{v_{\cal O}}} \right)
\end{array}
\end{equation}

Due to ${{q^\rho }\left( {{z^\rho }\left| X \right.} \right)}$ and ${{q^v }\left( {{z^v }\left| X \right.} \right)}$ being a normal distribution, the specific objective function can be derived:

\begin{equation}
\begin{array}{l}
los{s_{\rho VAE}}\left( {{\phi ^\rho },{\theta ^\rho }} \right) = \frac{1}{{{N_{\cal O}}}}\sum\limits_{i = 1}^{{N_{\cal O}}} {{{\left( {\hat \rho _{\cal O}^i - \rho _{\cal O}^i} \right)}^2}}  - \frac{1}{2}\sum\limits_{d = 1}^D {\left( {\log \sigma {{_d^\rho }^2} - \mu {{_d^\rho }^2} - \sigma {{_d^\rho }^2} + 1} \right)} \\
los{s_{vVAE}}\left( {{\phi ^v},{\theta ^v}} \right) = \frac{1}{{{N_{\cal O}}}}\sum\limits_{i = 1}^{{N_{\cal O}}} {{{\left( {\hat v_{\cal O}^i - v_{\cal O}^i} \right)}^2}}  - \frac{1}{2}\sum\limits_{d = 1}^D {\left( {\log \sigma {{_d^v}^2} - \mu {{_d^v}^2} - \sigma {{_d^v}^2} + 1} \right)} 
\end{array}
\end{equation}
where $D$ is the dimension of latent space ${z^\rho },{z^v}$, $d$ means each component of the encoder moments $\mu _z^\rho ,\sigma _z^\rho ,\mu _z^v,\sigma _z^v$.

\subsubsection{Stochastic processes used for regularization training}
By using the stochastic fundamental diagram (speed-density relationship) modeled in Section \ref{sp}, traffic speed distribution at a given density can be determined. Therefore, we perform latent spatial sampling on the collocation points $\left( {x,t} \right) \in {\cal C}$  and generate the speed distribution ${\tilde v_{\cal C}}$. When we input a $\rho _{\cal C}^i$ to stochastic fundamental diagram, the physical distribution $\tilde v_{phy}^i$ can be obtained through Eq. \ref{pdf}. Thus, the deviation between the speeds distributions at selected densities using the stochastic FD model and the VAE network can be measured by KL divergence:

\begin{equation}
Los{s_{phy}} = \frac{1}{{{N_{\cal C}}}}\sum\limits_{i = 1}^{{N_{\cal C}}} {{D_{KL}}\left( {\tilde v_{\cal C}^i\left\| {\tilde v_{phy}^i} \right.} \right)}  = \frac{1}{{{N_{\cal C}}}}\sum\limits_{i = 1}^{{N_{\cal C}}} {\tilde v_{\cal C}^i \cdot } \log \frac{{\tilde v_{\cal C}^i}}{{\tilde v_{phy}^i}} = \frac{1}{{{N_{\cal C}}}}\sum\limits_{i = 1}^{{N_{\cal C}}} {\tilde v_{\cal C}^i \cdot } \log \frac{{\tilde v_{\cal C}^i}}{{\left( {\frac{1}{{{v_{max}}}}} \right)\frac{{{{\left( {\frac{{v_{\cal C}^i}}{{{v_{max}}}}} \right)}^{\alpha \left( {\rho _{\cal C}^i} \right) - 1}}{{\left( {1 - \frac{{v_{\cal C}^i}}{{{v_{max}}}}} \right)}^{\beta \left( {\rho _{\cal C}^i} \right) - 1}}}}{{B\left( {\alpha \left( {\rho _{\cal C}^i} \right),\beta \left( {\rho _{\cal C}^i} \right)} \right)}}}}
\end{equation}
where ${{N_{\cal C}}}$ is total number of collocation samples, consistent with Eq. \ref{phylwr} and Eq. \ref{phyarz}.  

\subsubsection{Combined loss function}
As shown in Fig. \ref{planB}, the entire model is optimized from both data-driven and model-based perspectives. The total loss function is combined with three parts: data loss caused by density VAE network, data loss caused by speed VAE network, and physical loss, defined as follows:

\begin{equation}
Los{s_{Total}}\left( \varphi  \right) = {\kappa _1}Los{s_{\rho VAE}} + {\kappa _2}Los{s_{vVAE}} + {\kappa _3}Los{s_{phy}}
\end{equation}
where $\varphi $ represent the set of hyperparameters, ${\kappa _1}$, ${\kappa _2}$ and ${\kappa _3}$ are weight coefficients. 

Given the training data, our goal is to solve ${\varphi ^ * } = argminLos{s_{Total}}\left( \varphi  \right)$. Finally, the
${\varphi ^ * }$-parameterized network can be applied to approximate the
global traffic states on $\left( {x,t} \right) \in {\cal G}$.

\section{Experiments}\label{s4}
\subsection{Data description}
\begin{figure}[pos=htbp]
    \centering
\includegraphics[width=0.9\textwidth]{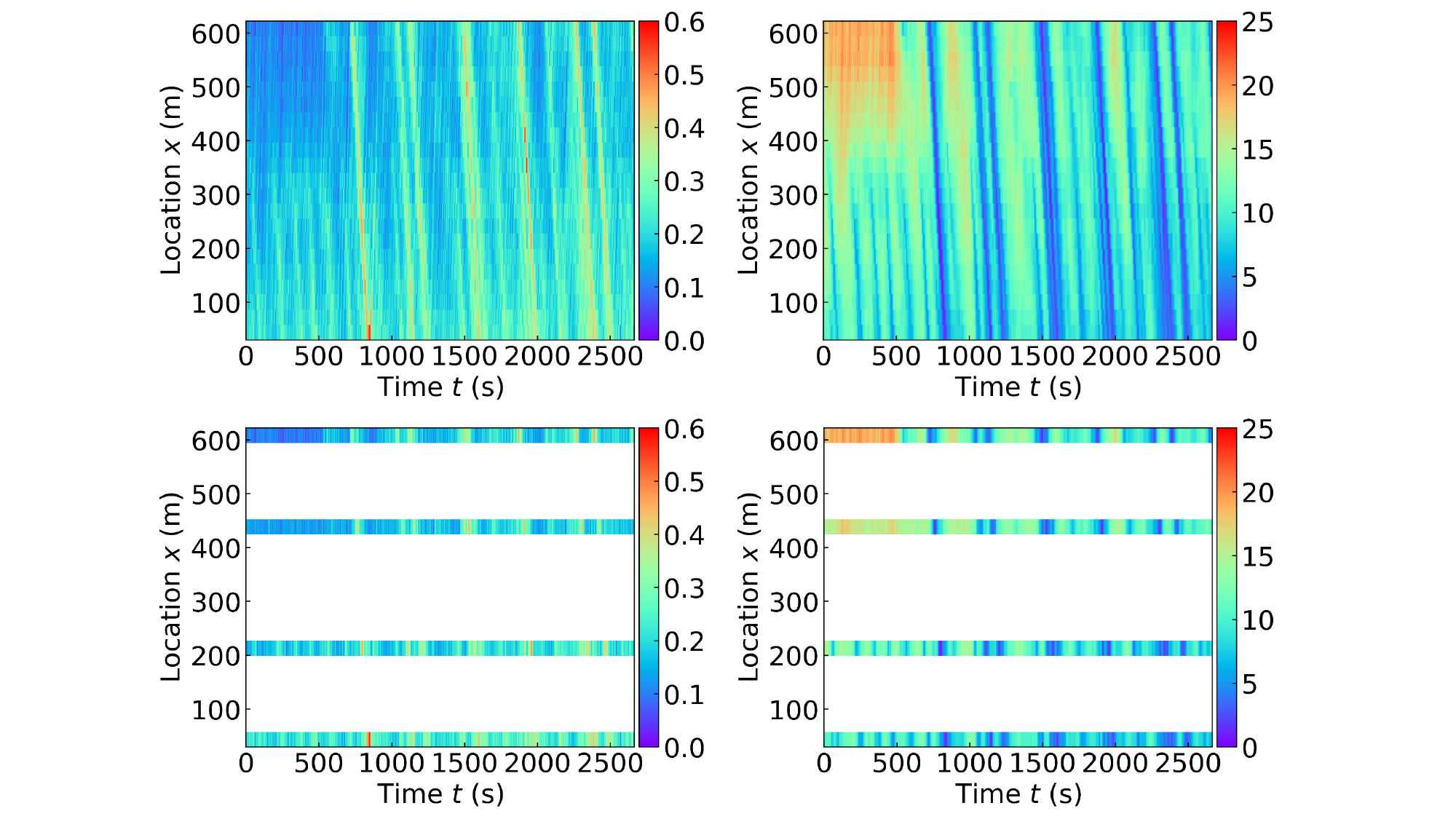}
    \caption{\centering{Experimental dataset}}
	\label{data}
\end{figure}

The real-world Next Generation Simulation (NGSIM) dataset was used to verify the effectiveness of the proposed TSE method in this paper. Within a monitoring area of approximately 680 meters and 2770 seconds, the position and movements of each vehicle are converted from camera video. This dataset has been widely used in TSE works \citep{sun2017simultaneous, wu2024traffic}. We use NGSIM dataset to generate simulated fixed detector data \citep{shi2021physics}. Specifically, a distance of 30 meters and a time interval of 1.5 seconds was selected, during which we counted all vehicles passing through the detector to obtain the flow, and averaged the instantaneous speed of the passing vehicles to obtain the speed. It should be noted that density can also be calculated, and the target estimated state variables in this paper are velocity and density. After processing, valid grids with a spatiotemporal dimension of 21X1770 were finally obtained, representing 21 loop detectors and 1770 time steps. Fig. \ref{data} shows the global state and a schematic of sparse detection data scenarios, with four virtual detectors.

\subsection{Experimental setup}
The mean absolute error (MAE), root mean square error (RMSE), and  ${{L^2}}$ relative error are selected to evaluate the performance of proposed SPIDL models, and the estimated uncertainty is reflected through confidence intervals (CI). All models are executed on an Intel(R) Xeon(R) Platinum 8255C CPU @2.50GHz and an NVIDIA RTX 3080 GPU, using Pytorch framework. We set 4, 6, 8, 10 and 12 as the number of loop detectors in sparse detection data scenarios. The hyperparameters for \text{$\alpha$}-SPIDL and \text{$\cal B$}-SPIDL are detailed in Table \ref{parameters}. 

\begin{table}[pos=h]
  \centering
  \caption{Parameters of SPIDL}
    \begin{tabular}{cccc}
    \toprule
    Model & Type  & Hyperparameter & Values \\
    \midrule
    \multirow{5}[4]{*}{\text{$\alpha$}-SPIDL} & Physics & Percentage number $m$ & 15  (see Table \ref{Calibration} for details) \\
\cmidrule{2-4}          & \multirow{4}[2]{*}{Neural network} & Learning rate & 0.001 \\
          &       & Layers & 11 \\
          &       & Hidden nodes & 64 \\
          &       & Optimizer & Adam \\
    \midrule
    \multirow{9}[4]{*}{\text{$\cal B$}-SPIDL} & Physics & Interval number & 30 \\
\cmidrule{2-4}          & \multirow{8}[2]{*}{VAE network} & Learning rate & 0.0005 \\
          &       & Encoder layers $L$ & 4 \\
          &       & Encoder hidden nodes & 32 \\
          &       & Latent dimension $D$ & 32 \\
          &       & Noisy ${\delta ^\rho },{\delta ^v}$ & Randn\_like(std) \\
          &       & Decoder layers $M$ & 4 \\
          &       & Decoder hidden nodes & 32 \\
          &       & Optimizer & Adam \\
    \bottomrule
    \end{tabular}%
  \label{parameters}%
\end{table}%

\subsection{Results and discussion}
\subsubsection{Accuracy index}
\begin{figure}[pos=htbp]
    \centering
\includegraphics[width=1\textwidth]{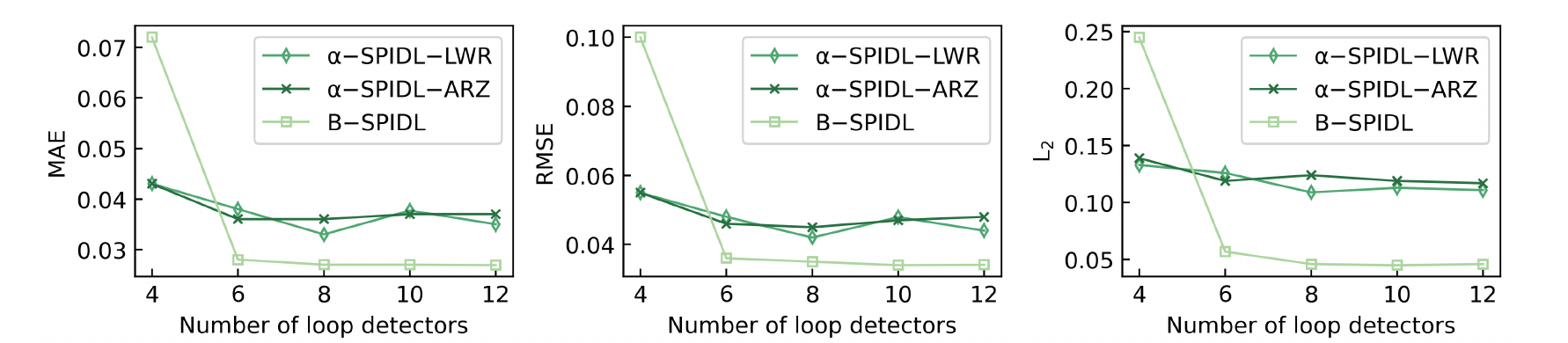}
    \caption{\centering{Evolution of density estimation performance with different numbers of loop detectors}}
	\label{Index_density}
\end{figure}

\begin{figure}[pos=htbp]
    \centering
\includegraphics[width=1\textwidth]{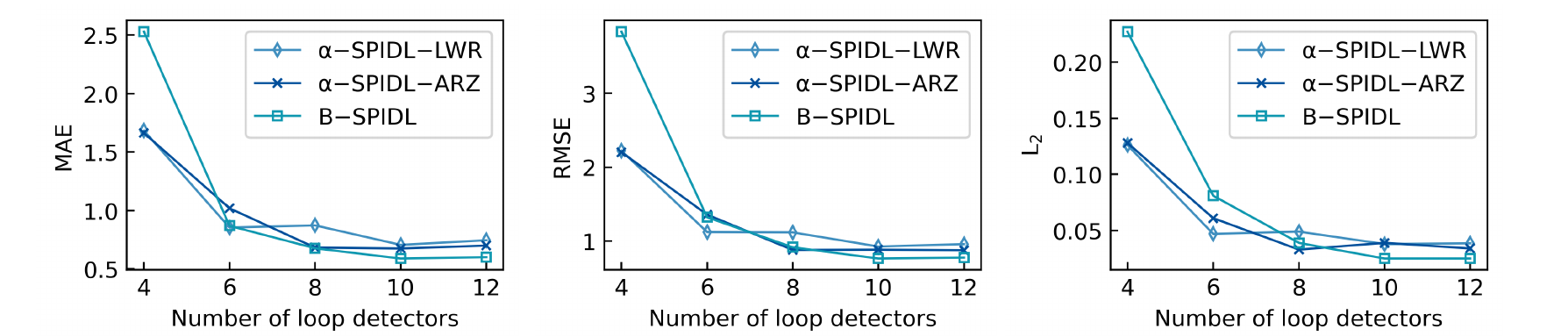}
    \caption{\centering{Evolution of speed estimation performance with different numbers of loop detectors}}
	\label{Index_speed}
\end{figure}

Fig. \ref{Index_density} and Fig. \ref{Index_speed}  correspond to the accuracy metrics of density and speed estimation for \text{$\alpha$}-SPIDL-LWR, \text{$\alpha$}-SPIDL-ARZ and \text{$\cal B$}-SPIDL proposed in this paper. The horizontal axis of all subgraphs represents the number of loop detectors, and the vertical axis represents the accuracy metrics. In terms of density estimation in Fig. \ref{Index_density}, when there are only 4 loops (20\% available detection data), the MAE, RMSE, and ${{L^2}}$ of \text{$\cal B$}-SPIDL are 2.528, 3.835, and 0.227, respectively, with lower accuracy than \text{$\alpha$}-SPIDL-LWR and \text{$\alpha$}-SPIDL-ARZ. As the number of loops increased to 6, there was a marked enhancement in the precision of all proposed models. This improvement can be attributed to the provision of richer data enabling the network to better learn and adapt to the intrinsic features of the data. Furthermore, increasing the number of loops has a limited effect on improving the accuracy, and the performance tends to stabilize, achieving high accuracy. It should be noted that 10 loops are only about 50\% of the total working coils, which is completely acceptable compared to the effective working rate of about 70\% in practical applications. An additional interesting finding is that the performance of \text{$\cal B$}-SPIDL gradually surpasses that of \text{$\alpha$}-SPIDL as the loops increase. This may be due to the "less is more" design of \text{$\cal B$}-SPIDL, the advantages of data mechanism discovery are gradually revealed. Moving to the speed estimation in Fig. \ref{Index_speed}, the accuracy evolution trend in each sparse detection data scenario is not completely consistent with the density estimation, but the overall trend corresponds. This makes sense because the proposed SPLID models is a multi-task (speed and density) estimation, and the two variables inevitably affect each other. Of course, with the increase of detection data, the performance of \text{$\cal B$}-SPIDL gradually surpasses that of \text{$\alpha$}-SPIDL, which is also reflected in speed estimation. Overall, the proposed SPILD models can accurately estimate the global traffic flow state based on sparse data, with MAE, RMSE, and L2 only around 0.5, 0.7, and 0.025, respectively.

\subsubsection{Comparison between the estimated state heatmap and the observed state heatmap}
\begin{figure}[pos=htbp]
    \centering
\includegraphics[width=1\textwidth]{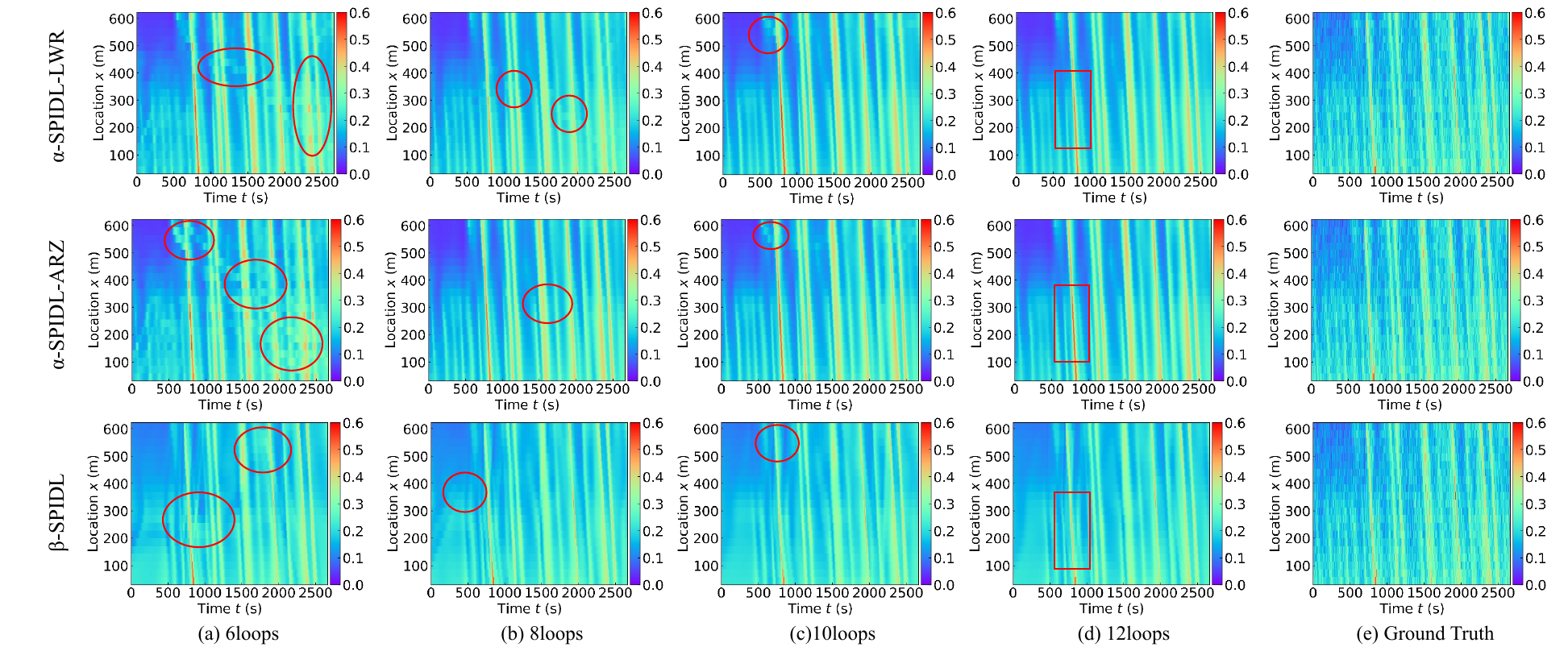}
    \caption{\centering{The estimated density heatmap of proposed SPIDL models vs ground truth}}
	\label{densityHeatmap}
\end{figure}

\begin{figure}[pos=htbp]
    \centering
\includegraphics[width=1\textwidth]{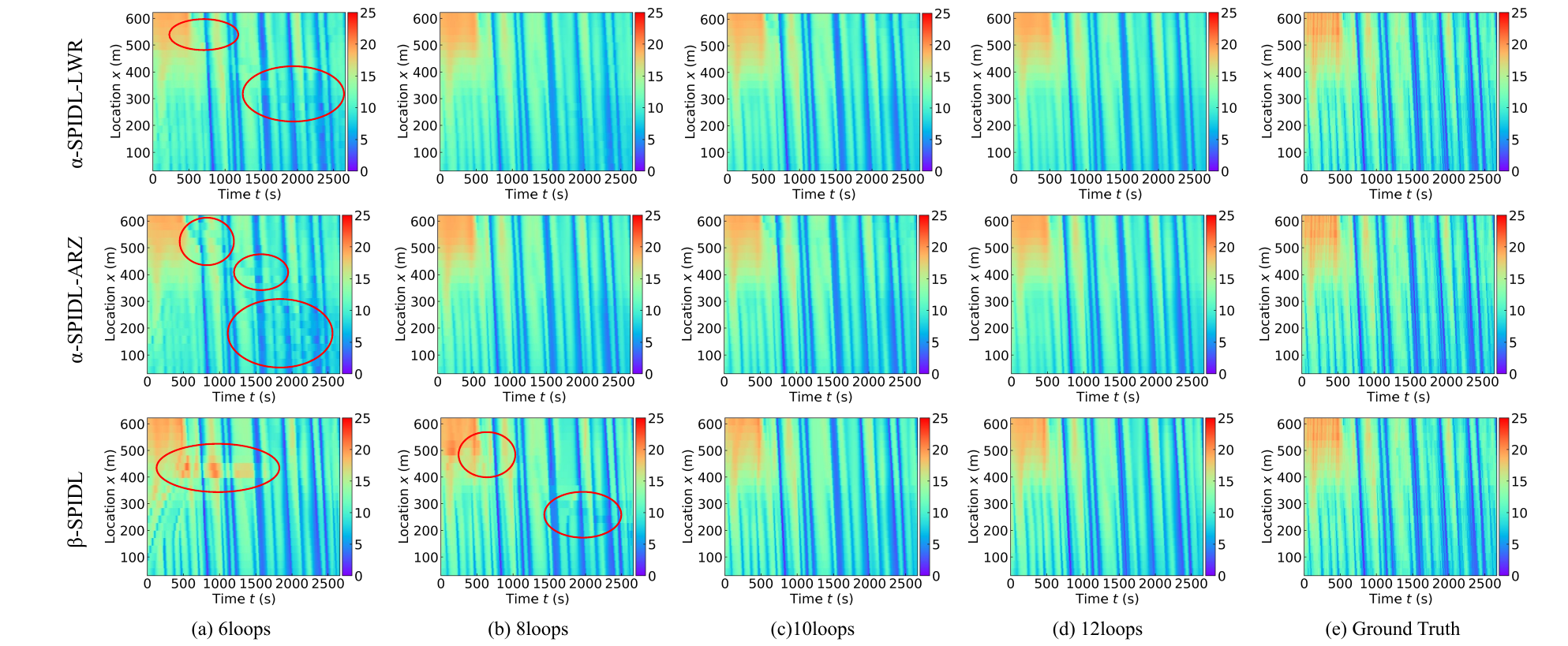}
    \caption{\centering{The estimated speed heatmap of proposed SPIDL models vs ground truth}}
	\label{speedHeatmap}
\end{figure}

The spatiotemporal heatmaps of density and speed estimated by the proposed SPIDL models are intuitively presented in Fig. \ref{densityHeatmap} and Fig. \ref{speedHeatmap}, respectively, corresponding to Fig. \ref{Index_density} and Fig. \ref{Index_speed}. The horizontal comparison of the two sets of figures corresponds to different numbers of loop detectors, specifically 6, 8, 10, and 12, while the vertical comparison showcases \text{$\alpha$}-SPIDL-LWR, \text{$\alpha$}-SPIDL-ARZ and \text{$\cal B$}-SPIDL. The rightmost subplot in each figure presents the ground truth traffic state for the entire road segment, where congestion and typical stop-and-go waves can be observed.

In Fig. \ref{densityHeatmap}, the color bar transitions from blue to red, representing the range from high-density congested flow to free flow. With 6 detectors, the estimation results are not entirely accurate but generally capture the overall trend of traffic evolution. With 8 detectors, the estimations obtained from all three SPIDL models closely match the ground truth. They accurately capture the stable traffic characteristics during the first 1000 seconds and the subsequent shock waves, clearly depicting the backward propagation of congestion. Notably, by comparing the estimation results in Fig. \ref{densityHeatmap} (d), as highlighted by the red box, it is evident that \text{$\cal B$}-SPIDL does not overestimate the shock waves, whereas \text{$\alpha$}-SPIDL-LWR and \text{$\alpha$}-SPIDL-ARZ fail to capture this aspect. This observation aligns with the findings from Fig. \ref{Index_density}, which demonstrate the advantage of \text{$\cal B$}-SPIDL over \text{$\alpha$}-SPIDL as more data is available. Overall, as expected, all proposed SPIDL models demonstrate effective and accurate traffic state estimation capabilities under sparse detection data scenarios. Continuing to observe speed estimation, firstly, as marked by the red circle in Fig.\ref{speedHeatmap} (a), there is still considerable room for improvement in the estimation accuracy of the SPIDL models. Subsequently, Fig.\ref{speedHeatmap} (b), which increased the number of loops to 8, reproduces a more realistic depiction of the actual traffic flow evolution. Furthermore, the spatiotemporal speed evolution presented in Fig.\ref{speedHeatmap} (c) and Fig.\ref{speedHeatmap} (d) is highly consistent with the actual observations. This further confirms that, driven by the fusion of stochastic fundamental diagrams and deep learning, SPIDL models can achieve precise and meticulous traffic feature capture.

\subsubsection{Comparison between the state estimation of \text{$\alpha$}-SPIDL models and observation at some locations}
\begin{figure}[pos=htbp]
    \centering
\includegraphics[width=1\textwidth]{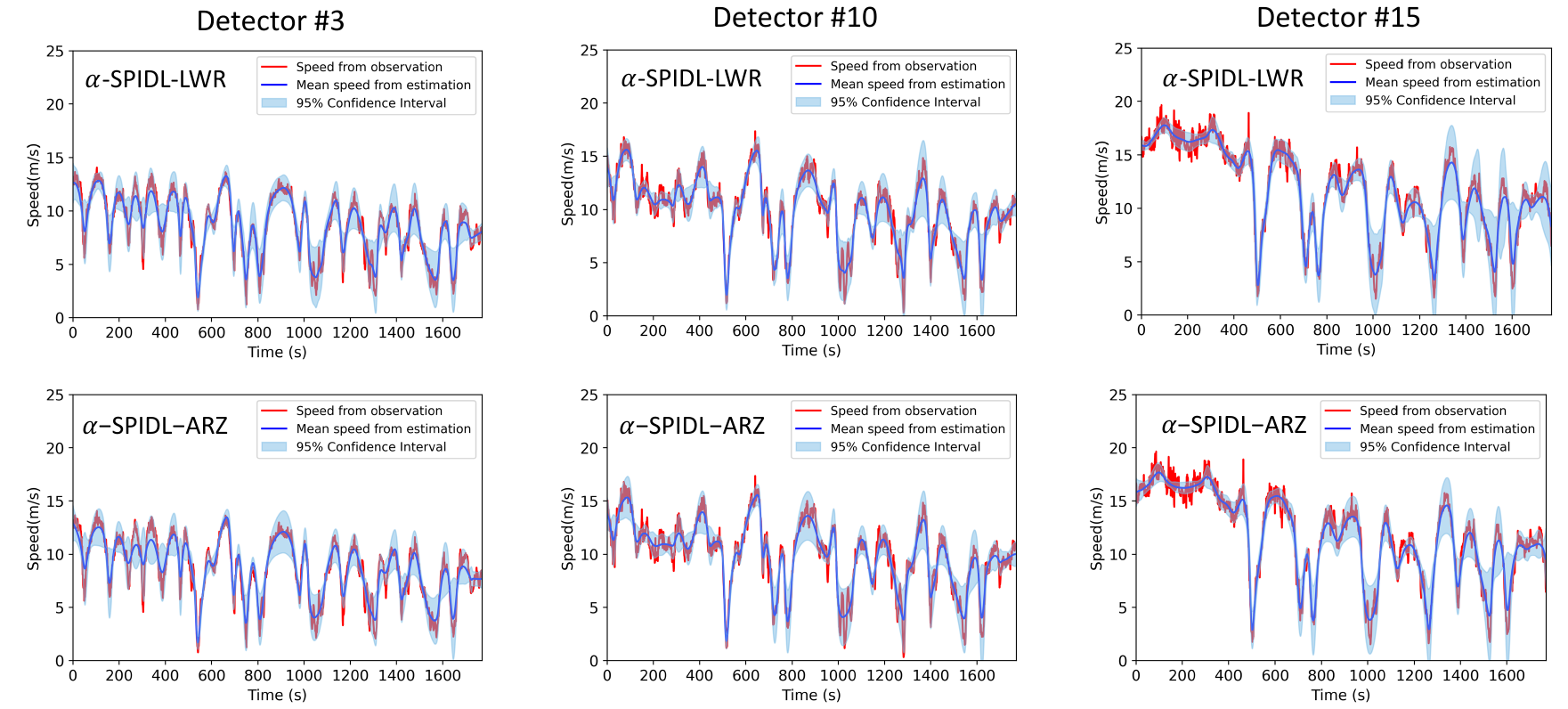}
    \caption{\centering{Stochastic speed estimation by \text{$\alpha$}-SPIDL models vs observation at some locations}}
	\label{Evolution}
\end{figure}

Fig. \ref{Evolution} depicts the temporal dynamics of estimated speed against the ground truth at detectors 3, 10 and 15, which is the "cross-section" profile of Fig. \ref{speedHeatmap}. The selected locations are representative and present different traffic evolution patterns. The selected sparse data scenario is 10 loops numbers, as the performance of \text{$\alpha$}-SPIDL-LWR and \text{$\alpha$}-SPIDL-ARZ tends to stabilize at 10, i.e. "converge". In Fig. \ref{Evolution}, the blue area between the upper and lower bound describes the trailing speed at 95\% confidence interval (CI). The red solid lines represent the detected speed, and the blue solid line is the average estimated speed. Fig. \ref{Evolution} shows that within the 95\% confidence interval, the detected speed fits well to the range of the stochastic estimation by \text{$\alpha$}-SPIDL-LWR and \text{$\alpha$}-SPIDL-ARZ, where it belongs within the upper and lower bound of the estimated speed. The overall observed evolution falls within the confidence region. Satisfactorily, the \text{$\alpha$}-SPIDL-LWR and \text{$\alpha$}-SPIDL-ARZ successfully replicated the stop-and-go waves that occurred due to the on-ramp flowing into a highly congested traffic situation, and the accuracy and robustness of proposed \text{$\alpha$}-SPIDL models are evident. 

\subsubsection{Estimated fundamental diagram of \text{$\cal B$}-SPIDL with confidence interval}
\begin{figure}[pos=htbp]
    \centering
\includegraphics[width=1\textwidth]{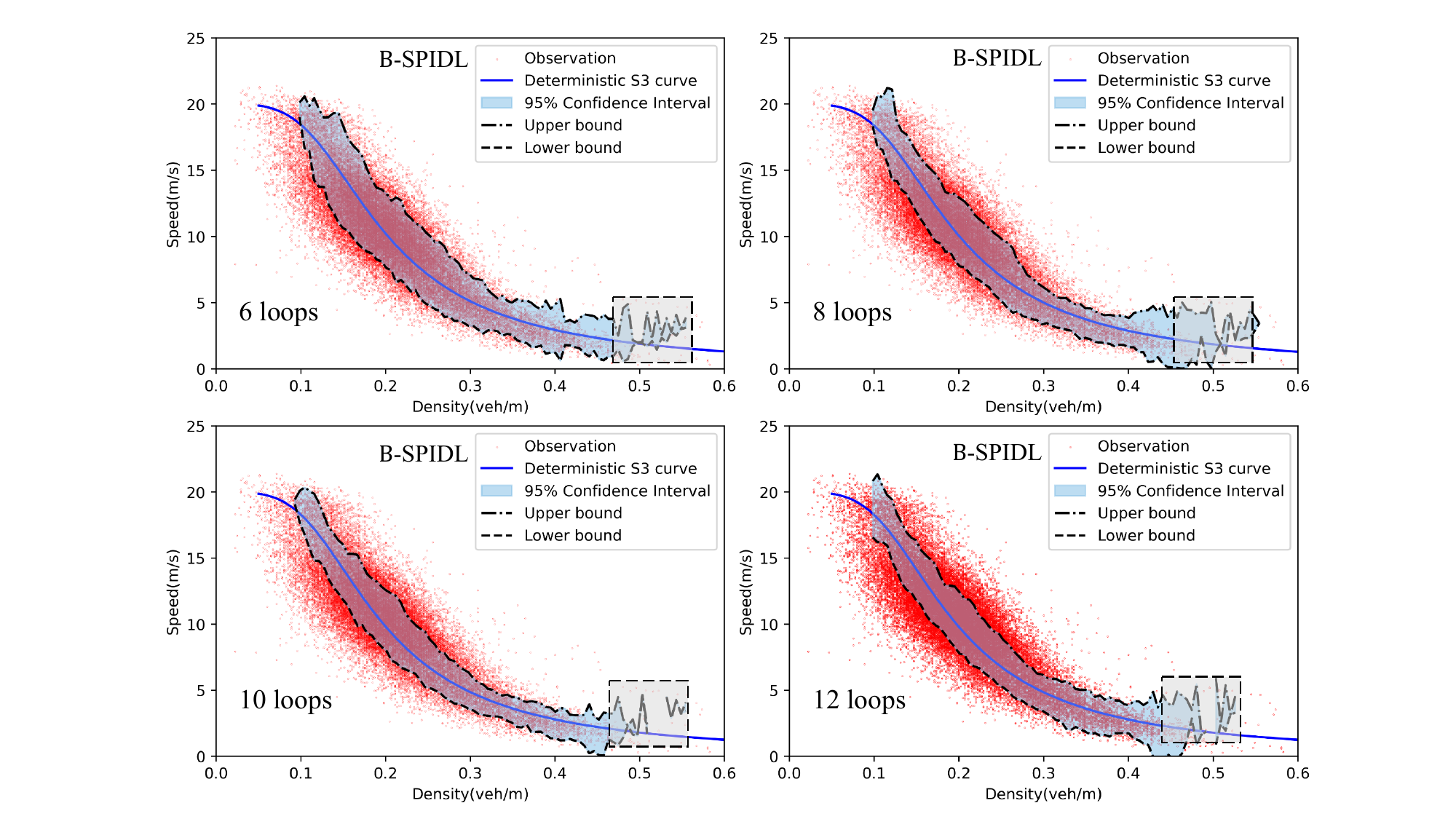}
    \caption{\centering{Observation scatters vs estimated fundamental diagram by \text{$\cal B$}-SPIDL with confidence interval}}
	\label{FD_CI}
\end{figure}

After confirming the effectiveness of \text{$\alpha$}-SPIDL models, we examine \text{$\cal B$}-SPIDL from another perspective. Fig.\ref{FD_CI} shows the observed versus estimated scatters of the speed–density relationship under different degrees of sparse data, across 6 loops, 8 loops, 10loos and 12 loops, respectively. The black dotted line represents the upper and lower bounds, the red dot represents the observation scatter, the blue area corresponds to the 95\% confidence interval, and the blue solid line represents the curve fitted with the S3 model \citep{cheng2021s}. The results demonstrated accurate approximations of the generated scatters compared to the true values. In other words, the estimated fundamental diagram with 95\% CI by \text{$\cal B$}-SPIDL approximates the observation well. It should be noted that, as indicated by the grey rectangle in Fig. \ref{FD_CI}, there is a noticeable fluctuation at the tail. This makes sense because the traffic flow has completely transitioned to congested flow, which once again confirms that \text{$\cal B$}-SPIDL can capture the scattering effect in the speed-density relationship and achieve stochastic traffic state estimation.

\subsubsection{The probability distribution of speeds with varying density levels estimated by SPIDL models}
\begin{figure}[pos=htbp]
    \centering
\includegraphics[width=1\textwidth]{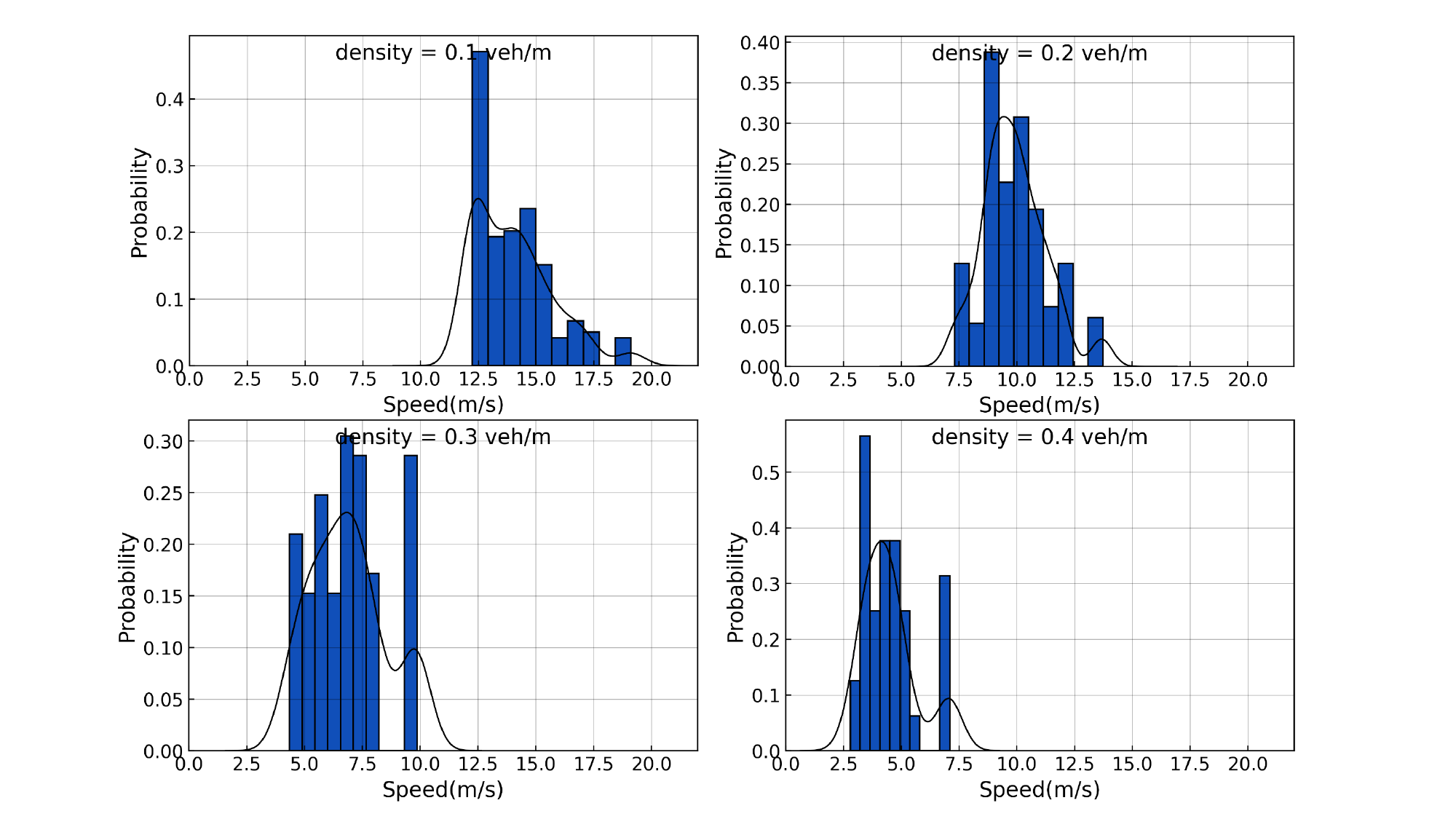}
    \caption{\centering{Speed distribution at different traffic densities using \text{$\alpha$}-SPIDL-LWR}}
	\label{hista1}
\end{figure}

\begin{figure}[pos=htbp]
    \centering
\includegraphics[width=1\textwidth]{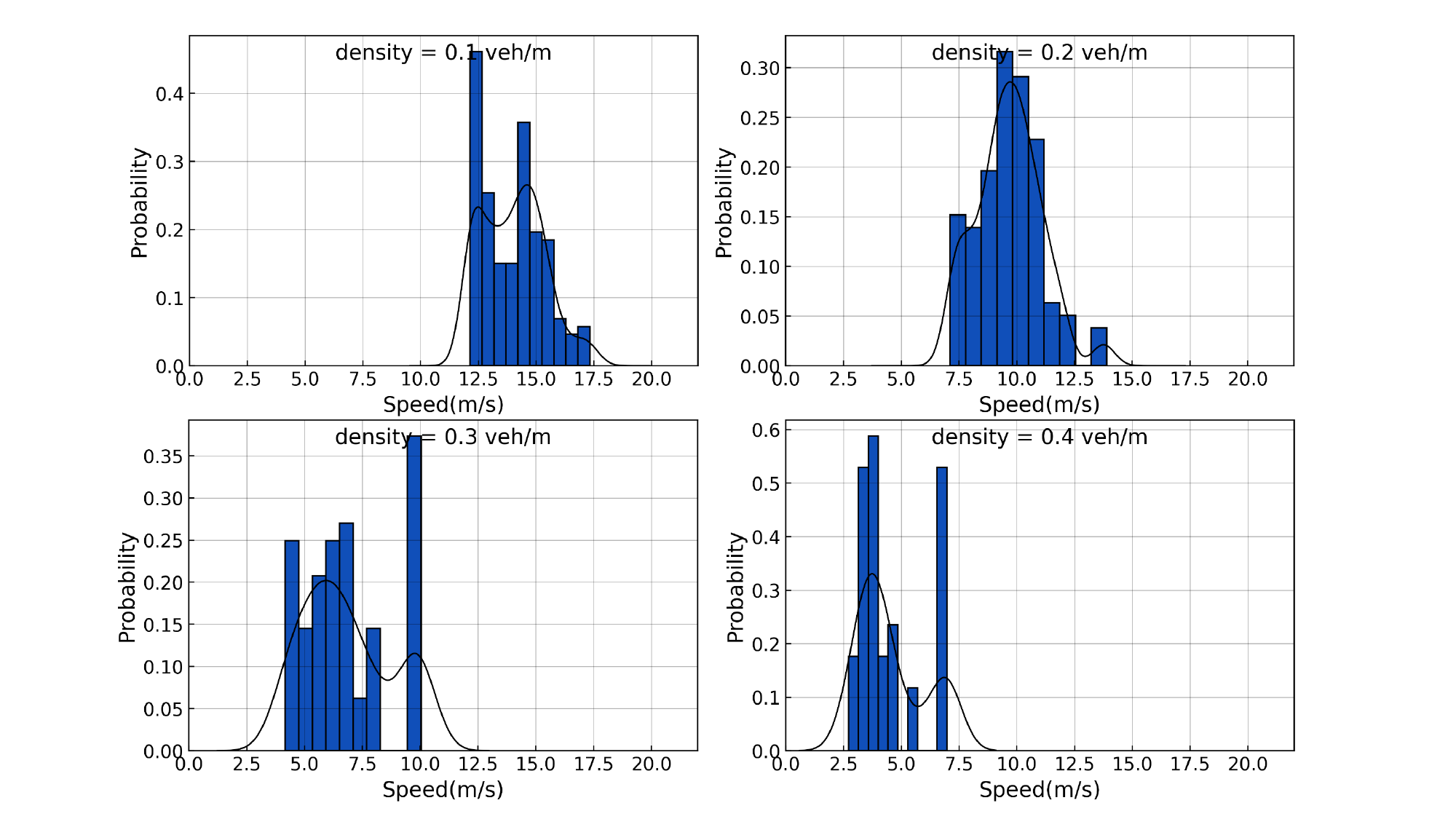}
    \caption{\centering{Speed distribution at different traffic densities using \text{$\alpha$}-SPIDL-ARZ}}
	\label{hista2}
\end{figure}

\begin{figure}[pos=htbp]
    \centering
\includegraphics[width=1\textwidth]{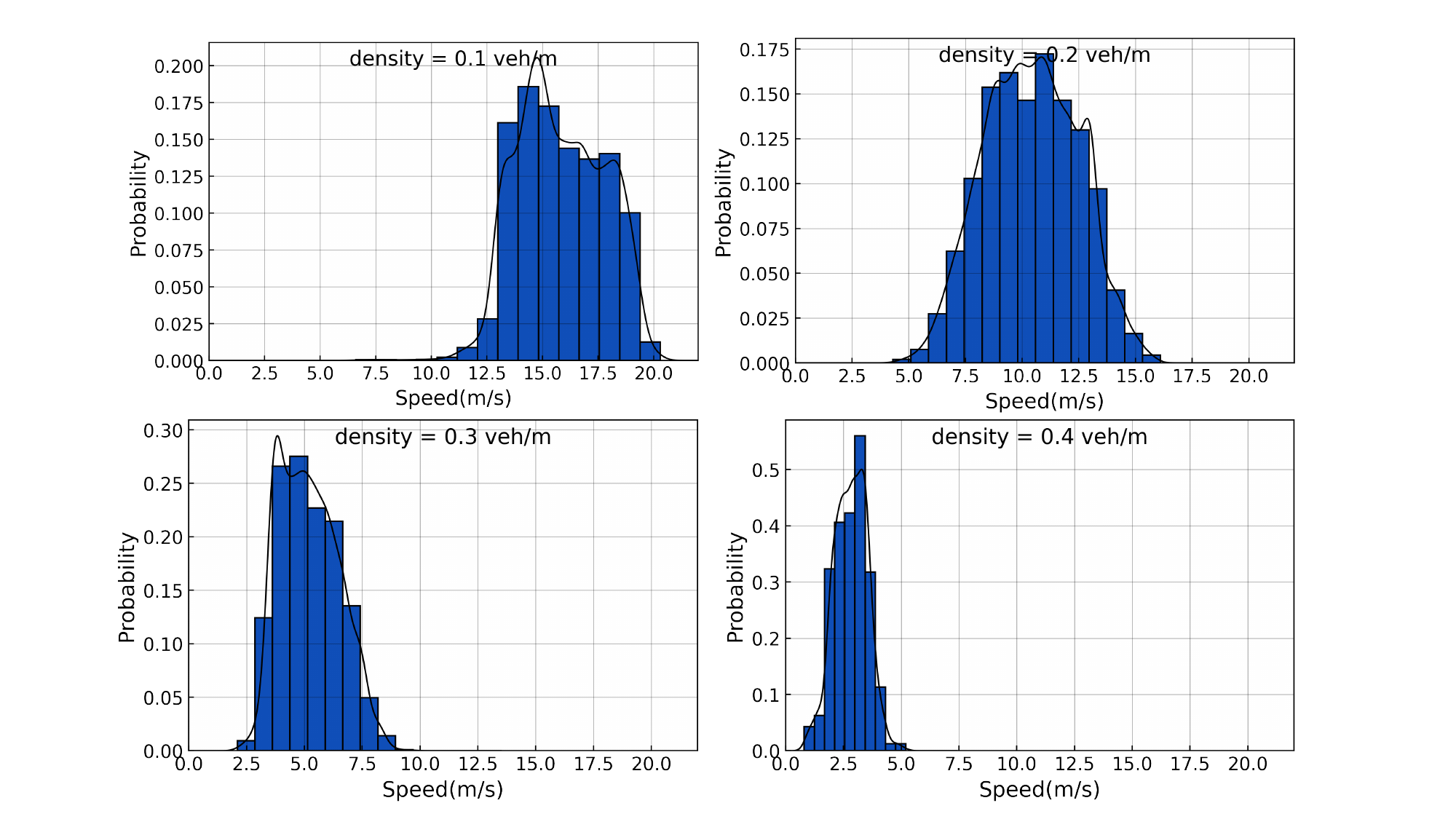}
    \caption{\centering{Speed distribution at different traffic densities using ${\cal B} - SPIDL$}}
	\label{histb}
\end{figure}

Fig. \ref{hista1}, Fig. \ref{hista2} and Fig. \ref{histb} respectively show the probability distributions of speed estimated using \text{$\alpha$}-SPIDL-LWR, \text{$\alpha$}-SPIDL-ARZ, and ${\cal B}$-SPIDL at the selected traffic density. The selected given densities are 0.1veh/m, 0.2veh/m, 0.3veh/m, and 0.4veh/m, respectively. Starting from the overall perspective, the estimated distribution effectively reflects the true distribution of real-world traffic flow, although there are differences between our proposed models. The trend presented in Fig. \ref{hista1}, Fig. \ref{hista2} and Fig. \ref{histb} can be summarized into two points: 1) the probability distribution of speeds at lower densities is spread over a wide range of speeds, whereas the distribution becomes narrower with the increase in traffic density; 2) Given a smaller density, the speed distribution is maintained at a larger speed value. As the given density increases, the speed distribution becomes smaller, which is consistent with the fundamental diagram estimated in Fig. \ref{FD_CI} earlier. For example, in Fig. \ref{histb}, when the density is 0.1veh/m, the distribution range of the speed estimated by ${\cal B}$-SPIDL is from 10m/s to 20m/s. When the density is 0.4veh/m, the speed distribution range is from 0m/s to 5m/s. Considering these, our research emphasizes the importance of fusing stochastic fundamental diagrams into PIDL architectures for traffic state estimation, which can accurately capture the uncertainty of traffic waves in traffic state estimation. In contrast, the deterministic models in existing PIDL-based TSE models yield one value of speed at a given density, which is unlikely to estimate the prevalent dynamical randomness effects that have been observed empirically.

\section{Conclusion} \label{s5}
In this paper, we aim to address the limitations caused by the physical knowledge being a solely deterministic model in the current PIDL-based TSE architectures. The deterministic model imposes homogenized physical constraints on the training of PIDL, neglecting the prevalent dynamical randomness effects of traffic flow. Therefore, we propose percentile-based SPIDL (\text{$\alpha$}-SPIDL) and distribution-based SPIDL (\text{$\cal B$}-SPIDL) frameworks for traffic state estimation from the perspective of extending PIDL to SPIDL. In \text{$\alpha$}-SPIDL, we select different $\alpha$ values and derive a family of percentile-based speed-density curves based on the theorem of total probability as reported in \citep{qu2017stochastic}, and then replace the deterministic physical knowledge in conventional PIDL-based TSE framework to correspondingly estimate a family of speed and density states, thus achieving the distributional estimation. Compared to the sequential training paradigms of \text{$\alpha$}-SPIDL, \text{$\cal B$}-SPIDL seems "less is more". Within the proposed \text{$\cal B$}-SPIDL, we first design a VAE network with two pairs of encoder-decoders assigned speed and density to unsupervisely learn an approximation of the probability distribution of the observed traffic state data. Moreover, the stochastic fundamental diagram derived from the standard Beta distribution \citep{ni2018modeling} is utilized and enforced in the framework. The aim is to compare the deviation with the probability distribution generated by the VAE, and then use this as a physical loss to integrate with the data-driven loss of network, ultimately performing back-propagation and training using the total loss. The experimental results based on the real-world NGSIM dataset underscore the applicability and effectiveness of proposed SPIDL models in sparse detection data scenarios. \text{$\alpha$}-SPIDL-LWR, \text{$\alpha$}-SPIDL-ARZ and \text{$\cal B$}-SPIDL can achieve accurate and robust estimation, with the scattering effect being well captured.

In future work, we will apply multi-source data fusion \citep{trinh2024stochastic} into SPIDL framework to achieve a complete representation of traffic features, thereby achieving higher performance traffic state estimation. In addition, more problem-oriented cutting-edge technologies will also be attempted, including but not limited to incorporating stochasticity in physics-informed graph convolutional network (PIGCN) \citep{mo2024pi} to extend current research to the road network scale. Besides, more potential fusion architectures of physics knowledge and data are also worth exploring, such as physics-guided neural networks (PgNNs) and physics-encoded neural networks (PeNNs) \citep{faroughi2024physics}.

\section*{Acknowledgement}
This research was supported by the project of the National Key R\&D Program of China (No. 2018YFB1601301), the National Natural Science Foundation of China (No. 71961137006, NO. 52302441), the Science and Technology Commission of Shanghai Municipality (No. 22dz1207500), and the China Scholarship Council (No.202306260118).

\section*{Appendix}

Table \ref{Calibration} calibrates the speed-density model at different percentile values represented through parameter $\alpha $. 

\setcounter{table}{0}
\renewcommand{\thetable}{A\arabic{table}}
\begin{table}[pos=h]
  \centering
  \caption{Calibrated parameters of percentile based fundamental diagrams within \text{$\alpha$}-SPIDL}
    \begin{tabular}{ccc}
    \toprule
    Percentage $\alpha$ & Critical density ${{\rho _{cr}}}$ (veh/m) & Free flow speed ${{v_f}}$ (m/s) \\
    \midrule
    0.01  & 0.30   & 26.59 \\
    0.08  & 0.26  & 25.17 \\
    0.15  & 0.24  & 24.89 \\
    0.22  & 0.23  & 24.66 \\
    0.29  & 0.23  & 24.44 \\
    0.36  & 0.22  & 24.22 \\
    0.43  & 0.22  & 24.00 \\
    0.50  & 0.21  & 23.76 \\
    0.57  & 0.21  & 23.49 \\
    0.64  & 0.20  & 23.18 \\
    0.71  & 0.20  & 22.81 \\
    0.78  & 0.20  & 22.33 \\
    0.85  & 0.19  & 21.70 \\
    0.92  & 0.19  & 20.66 \\
    0.99  & 0.18  & 15.73 \\
    \bottomrule
    \end{tabular}%
  \label{Calibration}%
\end{table}%

The probability density function and parameter provided in the previous Section. \ref{sp} is used to fit the experimental data. Fig. \ref{beta} plot the fitted spectrum of empirical distributions of vehicle speed, the fitting quality is acceptable.

\setcounter{figure}{0}
\renewcommand{\thefigure}{A\arabic{figure}}
\begin{figure}[pos=htbp]
    \centering
\includegraphics[width=0.8\textwidth]{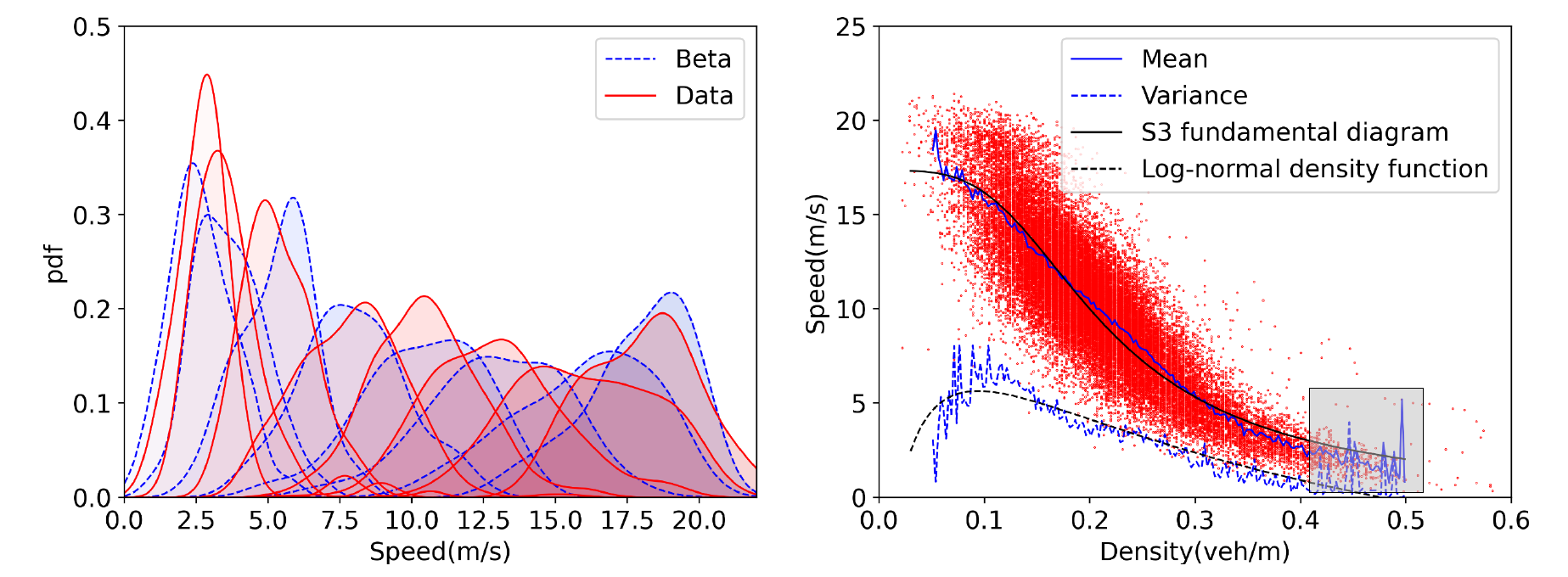}
    \caption{\centering{A spectrum of empirical distributions of vehicle speed fitted with Beta distribution.}}
	\label{beta}
\end{figure}

\bibliographystyle{cas-model2-names}

\bibliography{cas-sc-template}

\begin{thebibliography}{53}
\expandafter\ifx\csname natexlab\endcsname\relax\def\natexlab#1{#1}\fi
\providecommand{\url}[1]{\texttt{#1}}
\providecommand{\href}[2]{#2}
\providecommand{\path}[1]{#1}
\providecommand{\DOIprefix}{doi:}
\providecommand{\ArXivprefix}{arXiv:}
\providecommand{\URLprefix}{URL: }
\providecommand{\Pubmedprefix}{pmid:}
\providecommand{\doi}[1]{\href{http://dx.doi.org/#1}{\path{#1}}}
\providecommand{\Pubmed}[1]{\href{pmid:#1}{\path{#1}}}
\providecommand{\bibinfo}[2]{#2}
\ifx\xfnm\relax \def\xfnm[#1]{\unskip,\space#1}\fi
\bibitem[{Ahmed et~al.(2021)Ahmed, Ngoduy, Adnan and Baig}]{ahmed2021fundamental}
\bibinfo{author}{Ahmed, A.}, \bibinfo{author}{Ngoduy, D.}, \bibinfo{author}{Adnan, M.}, \bibinfo{author}{Baig, M.A.U.}, \bibinfo{year}{2021}.
\newblock \bibinfo{title}{On the fundamental diagram and driving behavior modeling of heterogeneous traffic flow using uav-based data}.
\newblock \bibinfo{journal}{Transportation research part A: policy and practice} \bibinfo{volume}{148}, \bibinfo{pages}{100--115}.
\bibitem[{Aw and Rascle(2000)}]{aw2000resurrection}
\bibinfo{author}{Aw, A.}, \bibinfo{author}{Rascle, M.}, \bibinfo{year}{2000}.
\newblock \bibinfo{title}{Resurrection of" second order" models of traffic flow}.
\newblock \bibinfo{journal}{SIAM journal on applied mathematics} \bibinfo{volume}{60}, \bibinfo{pages}{916--938}.
\bibitem[{Bai et~al.(2021)Bai, Wong, Xu, Chow and Lam}]{bai2021calibration}
\bibinfo{author}{Bai, L.}, \bibinfo{author}{Wong, S.}, \bibinfo{author}{Xu, P.}, \bibinfo{author}{Chow, A.H.}, \bibinfo{author}{Lam, W.H.}, \bibinfo{year}{2021}.
\newblock \bibinfo{title}{Calibration of stochastic link-based fundamental diagram with explicit consideration of speed heterogeneity}.
\newblock \bibinfo{journal}{Transportation Research Part B: Methodological} \bibinfo{volume}{150}, \bibinfo{pages}{524--539}.
\bibitem[{Boquet et~al.(2020)Boquet, Morell, Serrano and Vicario}]{boquet2020variational}
\bibinfo{author}{Boquet, G.}, \bibinfo{author}{Morell, A.}, \bibinfo{author}{Serrano, J.}, \bibinfo{author}{Vicario, J.L.}, \bibinfo{year}{2020}.
\newblock \bibinfo{title}{A variational autoencoder solution for road traffic forecasting systems: Missing data imputation, dimension reduction, model selection and anomaly detection}.
\newblock \bibinfo{journal}{Transportation Research Part C: Emerging Technologies} \bibinfo{volume}{115}, \bibinfo{pages}{102622}.
\bibitem[{Chen et~al.(2021)Chen, Lei, Saunier and Sun}]{chen2021low}
\bibinfo{author}{Chen, X.}, \bibinfo{author}{Lei, M.}, \bibinfo{author}{Saunier, N.}, \bibinfo{author}{Sun, L.}, \bibinfo{year}{2021}.
\newblock \bibinfo{title}{Low-rank autoregressive tensor completion for spatiotemporal traffic data imputation}.
\newblock \bibinfo{journal}{IEEE Transactions on Intelligent Transportation Systems} \bibinfo{volume}{23}, \bibinfo{pages}{12301--12310}.
\bibitem[{Cheng et~al.(2024)Cheng, Lin, Zhou and Liu}]{cheng2024analytical}
\bibinfo{author}{Cheng, Q.}, \bibinfo{author}{Lin, Y.}, \bibinfo{author}{Zhou, X.S.}, \bibinfo{author}{Liu, Z.}, \bibinfo{year}{2024}.
\newblock \bibinfo{title}{Analytical formulation for explaining the variations in traffic states: A fundamental diagram modeling perspective with stochastic parameters}.
\newblock \bibinfo{journal}{European Journal of Operational Research} \bibinfo{volume}{312}, \bibinfo{pages}{182--197}.
\bibitem[{Cheng et~al.(2021)Cheng, Liu, Lin and Zhou}]{cheng2021s}
\bibinfo{author}{Cheng, Q.}, \bibinfo{author}{Liu, Z.}, \bibinfo{author}{Lin, Y.}, \bibinfo{author}{Zhou, X.S.}, \bibinfo{year}{2021}.
\newblock \bibinfo{title}{An s-shaped three-parameter (s3) traffic stream model with consistent car following relationship}.
\newblock \bibinfo{journal}{Transportation Research Part B: Methodological} \bibinfo{volume}{153}, \bibinfo{pages}{246--271}.
\bibitem[{Daganzo(1994)}]{daganzo1994cell}
\bibinfo{author}{Daganzo, C.F.}, \bibinfo{year}{1994}.
\newblock \bibinfo{title}{The cell transmission model: A dynamic representation of highway traffic consistent with the hydrodynamic theory}.
\newblock \bibinfo{journal}{Transportation research part B: methodological} \bibinfo{volume}{28}, \bibinfo{pages}{269--287}.
\bibitem[{Deng et~al.(2013)Deng, Lei and Zhou}]{deng2013traffic}
\bibinfo{author}{Deng, W.}, \bibinfo{author}{Lei, H.}, \bibinfo{author}{Zhou, X.}, \bibinfo{year}{2013}.
\newblock \bibinfo{title}{Traffic state estimation and uncertainty quantification based on heterogeneous data sources: A three detector approach}.
\newblock \bibinfo{journal}{Transportation Research Part B: Methodological} \bibinfo{volume}{57}, \bibinfo{pages}{132--157}.
\bibitem[{Faroughi et~al.(2024)Faroughi, Pawar, Fernandes, Raissi, Das, Kalantari and Kourosh~Mahjour}]{faroughi2024physics}
\bibinfo{author}{Faroughi, S.A.}, \bibinfo{author}{Pawar, N.M.}, \bibinfo{author}{Fernandes, C.}, \bibinfo{author}{Raissi, M.}, \bibinfo{author}{Das, S.}, \bibinfo{author}{Kalantari, N.K.}, \bibinfo{author}{Kourosh~Mahjour, S.}, \bibinfo{year}{2024}.
\newblock \bibinfo{title}{Physics-guided, physics-informed, and physics-encoded neural networks and operators in scientific computing: Fluid and solid mechanics}.
\newblock \bibinfo{journal}{Journal of Computing and Information Science in Engineering} \bibinfo{volume}{24}, \bibinfo{pages}{040802}.
\bibitem[{Greenshields et~al.(1935)Greenshields, Bibbins, Channing and Miller}]{greenshields1935study}
\bibinfo{author}{Greenshields, B.D.}, \bibinfo{author}{Bibbins, J.R.}, \bibinfo{author}{Channing, W.}, \bibinfo{author}{Miller, H.H.}, \bibinfo{year}{1935}.
\newblock \bibinfo{title}{A study of traffic capacity}, in: \bibinfo{booktitle}{Highway research board proceedings}, \bibinfo{organization}{Washington, DC}. pp. \bibinfo{pages}{448--477}.
\bibitem[{Huang et~al.(2023a)Huang, Biswas and Agarwal}]{huang2023incorporating}
\bibinfo{author}{Huang, A.J.}, \bibinfo{author}{Biswas, A.}, \bibinfo{author}{Agarwal, S.}, \bibinfo{year}{2023}a.
\newblock \bibinfo{title}{Incorporating nonlocal traffic flow model in physics-informed neural networks}.
\newblock \bibinfo{journal}{arXiv preprint arXiv:2308.11818} .
\bibitem[{Huang et~al.(2023b)Huang, Yu, Chen and Lai}]{huang2023bridging}
\bibinfo{author}{Huang, H.}, \bibinfo{author}{Yu, J.}, \bibinfo{author}{Chen, J.}, \bibinfo{author}{Lai, R.}, \bibinfo{year}{2023}b.
\newblock \bibinfo{title}{Bridging mean-field games and normalizing flows with trajectory regularization}.
\newblock \bibinfo{journal}{Journal of Computational Physics} \bibinfo{volume}{487}, \bibinfo{pages}{112155}.
\bibitem[{Jabari and Liu(2013)}]{jabari2013stochastic}
\bibinfo{author}{Jabari, S.E.}, \bibinfo{author}{Liu, H.X.}, \bibinfo{year}{2013}.
\newblock \bibinfo{title}{A stochastic model of traffic flow: Gaussian approximation and estimation}.
\newblock \bibinfo{journal}{Transportation Research Part B: Methodological} \bibinfo{volume}{47}, \bibinfo{pages}{15--41}.
\bibitem[{Kurzhanskiy and Varaiya(2012)}]{kurzhanskiy2012guaranteed}
\bibinfo{author}{Kurzhanskiy, A.A.}, \bibinfo{author}{Varaiya, P.}, \bibinfo{year}{2012}.
\newblock \bibinfo{title}{Guaranteed prediction and estimation of the state of a road network}.
\newblock \bibinfo{journal}{Transportation research part C: emerging technologies} \bibinfo{volume}{21}, \bibinfo{pages}{163--180}.
\bibitem[{Laval et~al.(2012)Laval, He and Castrillon}]{laval2012stochastic}
\bibinfo{author}{Laval, J.A.}, \bibinfo{author}{He, Z.}, \bibinfo{author}{Castrillon, F.}, \bibinfo{year}{2012}.
\newblock \bibinfo{title}{Stochastic extension of newell's three-detector method}.
\newblock \bibinfo{journal}{Transportation research record} \bibinfo{volume}{2315}, \bibinfo{pages}{73--80}.
\bibitem[{Lei et~al.(2024)Lei, Xu and Wang}]{lei2024conditional}
\bibinfo{author}{Lei, D.}, \bibinfo{author}{Xu, M.}, \bibinfo{author}{Wang, S.}, \bibinfo{year}{2024}.
\newblock \bibinfo{title}{A conditional diffusion model for probabilistic estimation of traffic states at sensor-free locations}.
\newblock \bibinfo{journal}{Transportation Research Part C: Emerging Technologies} \bibinfo{volume}{166}, \bibinfo{pages}{104798}.
\bibitem[{Li et~al.(2024)Li, Li, Xu and Liu}]{li2024self}
\bibinfo{author}{Li, J.}, \bibinfo{author}{Li, R.}, \bibinfo{author}{Xu, L.}, \bibinfo{author}{Liu, J.}, \bibinfo{year}{2024}.
\newblock \bibinfo{title}{Self-supervised generative adversarial learning with conditional cyclical constraints towards missing traffic data imputation}.
\newblock \bibinfo{journal}{Knowledge-Based Systems} \bibinfo{volume}{284}, \bibinfo{pages}{111233}.
\bibitem[{Lighthill and Whitham(1955)}]{lighthill1955kinematic}
\bibinfo{author}{Lighthill, M.J.}, \bibinfo{author}{Whitham, G.B.}, \bibinfo{year}{1955}.
\newblock \bibinfo{title}{On kinematic waves ii. a theory of traffic flow on long crowded roads}.
\newblock \bibinfo{journal}{Proceedings of the royal society of london. series a. mathematical and physical sciences} \bibinfo{volume}{229}, \bibinfo{pages}{317--345}.
\bibitem[{Mo et~al.(2022a)Mo, Fu and Di}]{mo2022quantifying}
\bibinfo{author}{Mo, Z.}, \bibinfo{author}{Fu, Y.}, \bibinfo{author}{Di, X.}, \bibinfo{year}{2022}a.
\newblock \bibinfo{title}{Quantifying uncertainty in traffic state estimation using generative adversarial networks}, in: \bibinfo{booktitle}{2022 IEEE 25th International Conference on Intelligent Transportation Systems (ITSC)}, \bibinfo{organization}{IEEE}. pp. \bibinfo{pages}{2769--2774}.
\bibitem[{Mo et~al.(2024)Mo, Fu and Di}]{mo2024pi}
\bibinfo{author}{Mo, Z.}, \bibinfo{author}{Fu, Y.}, \bibinfo{author}{Di, X.}, \bibinfo{year}{2024}.
\newblock \bibinfo{title}{Pi-neugode: Physics-informed graph neural ordinary differential equations for spatiotemporal trajectory prediction}, in: \bibinfo{booktitle}{Proceedings of the 23rd International Conference on Autonomous Agents and Multiagent Systems}, pp. \bibinfo{pages}{1418--1426}.
\bibitem[{Mo et~al.(2022b)Mo, Fu, Xu and Di}]{mo2022trafficflowgan}
\bibinfo{author}{Mo, Z.}, \bibinfo{author}{Fu, Y.}, \bibinfo{author}{Xu, D.}, \bibinfo{author}{Di, X.}, \bibinfo{year}{2022}b.
\newblock \bibinfo{title}{Trafficflowgan: Physics-informed flow based generative adversarial network for uncertainty quantification}, in: \bibinfo{booktitle}{Joint European Conference on Machine Learning and Knowledge Discovery in Databases}, \bibinfo{organization}{Springer}. pp. \bibinfo{pages}{323--339}.
\bibitem[{Ngoduy(2008)}]{ngoduy2008applicable}
\bibinfo{author}{Ngoduy, D.}, \bibinfo{year}{2008}.
\newblock \bibinfo{title}{Applicable filtering framework for online multiclass freeway network estimation}.
\newblock \bibinfo{journal}{Physica A: Statistical Mechanics and its Applications} \bibinfo{volume}{387}, \bibinfo{pages}{599--616}.
\bibitem[{Ngoduy(2011a)}]{ngoduy2011low}
\bibinfo{author}{Ngoduy, D.}, \bibinfo{year}{2011}a.
\newblock \bibinfo{title}{Low-rank unscented kalman filter for freeway traffic estimation problems}.
\newblock \bibinfo{journal}{Transportation research record} \bibinfo{volume}{2260}, \bibinfo{pages}{113--122}.
\bibitem[{Ngoduy(2011b)}]{ngoduy2011multiclass}
\bibinfo{author}{Ngoduy, D.}, \bibinfo{year}{2011}b.
\newblock \bibinfo{title}{Multiclass first-order traffic model using stochastic fundamental diagrams}.
\newblock \bibinfo{journal}{Transportmetrica} \bibinfo{volume}{7}, \bibinfo{pages}{111--125}.
\bibitem[{Ngoduy(2021)}]{ngoduy2021noise}
\bibinfo{author}{Ngoduy, D.}, \bibinfo{year}{2021}.
\newblock \bibinfo{title}{Noise-induced instability of a class of stochastic higher order continuum traffic models}.
\newblock \bibinfo{journal}{Transportation Research Part B: Methodological} \bibinfo{volume}{150}, \bibinfo{pages}{260--278}.
\bibitem[{Ni et~al.(2018)Ni, Hsieh and Jiang}]{ni2018modeling}
\bibinfo{author}{Ni, D.}, \bibinfo{author}{Hsieh, H.K.}, \bibinfo{author}{Jiang, T.}, \bibinfo{year}{2018}.
\newblock \bibinfo{title}{Modeling phase diagrams as stochastic processes with application in vehicular traffic flow}.
\newblock \bibinfo{journal}{Applied Mathematical Modelling} \bibinfo{volume}{53}, \bibinfo{pages}{106--117}.
\bibitem[{Qu et~al.(2017)Qu, Zhang and Wang}]{qu2017stochastic}
\bibinfo{author}{Qu, X.}, \bibinfo{author}{Zhang, J.}, \bibinfo{author}{Wang, S.}, \bibinfo{year}{2017}.
\newblock \bibinfo{title}{On the stochastic fundamental diagram for freeway traffic: Model development, analytical properties, validation, and extensive applications}.
\newblock \bibinfo{journal}{Transportation research part B: methodological} \bibinfo{volume}{104}, \bibinfo{pages}{256--271}.
\bibitem[{Raissi(2018)}]{raissi2018deep}
\bibinfo{author}{Raissi, M.}, \bibinfo{year}{2018}.
\newblock \bibinfo{title}{Deep hidden physics models: Deep learning of nonlinear partial differential equations}.
\newblock \bibinfo{journal}{The Journal of Machine Learning Research} \bibinfo{volume}{19}, \bibinfo{pages}{932--955}.
\bibitem[{Shi et~al.(2021a)Shi, Mo and Di}]{shi2021physics}
\bibinfo{author}{Shi, R.}, \bibinfo{author}{Mo, Z.}, \bibinfo{author}{Di, X.}, \bibinfo{year}{2021}a.
\newblock \bibinfo{title}{Physics-informed deep learning for traffic state estimation: A hybrid paradigm informed by second-order traffic models}, in: \bibinfo{booktitle}{Proceedings of the AAAI Conference on Artificial Intelligence}, pp. \bibinfo{pages}{540--547}.
\bibitem[{Shi et~al.(2021b)Shi, Mo and Di}]{shi2021physicsA}
\bibinfo{author}{Shi, R.}, \bibinfo{author}{Mo, Z.}, \bibinfo{author}{Di, X.}, \bibinfo{year}{2021}b.
\newblock \bibinfo{title}{Physics-informed deep learning for traffic state estimation: A hybrid paradigm informed by second-order traffic models}, in: \bibinfo{booktitle}{Proceedings of the AAAI Conference on Artificial Intelligence}, pp. \bibinfo{pages}{540--547}.
\bibitem[{Shi et~al.(2023)Shi, Chen, Liu, Fan and Liang}]{shi2023physics}
\bibinfo{author}{Shi, Z.}, \bibinfo{author}{Chen, Y.}, \bibinfo{author}{Liu, J.}, \bibinfo{author}{Fan, D.}, \bibinfo{author}{Liang, C.}, \bibinfo{year}{2023}.
\newblock \bibinfo{title}{Physics-informed spatiotemporal learning framework for urban traffic state estimation}.
\newblock \bibinfo{journal}{Journal of Transportation Engineering, Part A: Systems} \bibinfo{volume}{149}, \bibinfo{pages}{04023056}.
\bibitem[{Sumalee et~al.(2011)Sumalee, Zhong, Pan and Szeto}]{sumalee2011stochastic}
\bibinfo{author}{Sumalee, A.}, \bibinfo{author}{Zhong, R.}, \bibinfo{author}{Pan, T.}, \bibinfo{author}{Szeto, W.}, \bibinfo{year}{2011}.
\newblock \bibinfo{title}{Stochastic cell transmission model (sctm): A stochastic dynamic traffic model for traffic state surveillance and assignment}.
\newblock \bibinfo{journal}{Transportation Research Part B: Methodological} \bibinfo{volume}{45}, \bibinfo{pages}{507--533}.
\bibitem[{Sun et~al.(2017)Sun, Jin and Ritchie}]{sun2017simultaneous}
\bibinfo{author}{Sun, Z.}, \bibinfo{author}{Jin, W.L.}, \bibinfo{author}{Ritchie, S.G.}, \bibinfo{year}{2017}.
\newblock \bibinfo{title}{Simultaneous estimation of states and parameters in newell’s simplified kinematic wave model with eulerian and lagrangian traffic data}.
\newblock \bibinfo{journal}{Transportation research part B: methodological} \bibinfo{volume}{104}, \bibinfo{pages}{106--122}.
\bibitem[{Tak et~al.(2016)Tak, Woo and Yeo}]{tak2016data}
\bibinfo{author}{Tak, S.}, \bibinfo{author}{Woo, S.}, \bibinfo{author}{Yeo, H.}, \bibinfo{year}{2016}.
\newblock \bibinfo{title}{Data-driven imputation method for traffic data in sectional units of road links}.
\newblock \bibinfo{journal}{IEEE Transactions on Intelligent Transportation Systems} \bibinfo{volume}{17}, \bibinfo{pages}{1762--1771}.
\bibitem[{Treiber et~al.(2011)Treiber, Kesting and Wilson}]{treiber2011reconstructing}
\bibinfo{author}{Treiber, M.}, \bibinfo{author}{Kesting, A.}, \bibinfo{author}{Wilson, R.E.}, \bibinfo{year}{2011}.
\newblock \bibinfo{title}{Reconstructing the traffic state by fusion of heterogeneous data}.
\newblock \bibinfo{journal}{Computer-Aided Civil and Infrastructure Engineering} \bibinfo{volume}{26}, \bibinfo{pages}{408--419}.
\bibitem[{Trinh et~al.(2024)Trinh, Keyvan-Ekbatani, Ngoduy and Robertson}]{trinh2024stochastic}
\bibinfo{author}{Trinh, X.S.}, \bibinfo{author}{Keyvan-Ekbatani, M.}, \bibinfo{author}{Ngoduy, D.}, \bibinfo{author}{Robertson, B.}, \bibinfo{year}{2024}.
\newblock \bibinfo{title}{Stochastic switching mode model based filters for urban arterial traffic estimation from multi-source data}.
\newblock \bibinfo{journal}{Transportation Research Part C: Emerging Technologies} \bibinfo{volume}{164}, \bibinfo{pages}{104664}.
\bibitem[{Trinh et~al.(2022)Trinh, Ngoduy, Keyvan-Ekbatani and Robertson}]{trinh2022incremental}
\bibinfo{author}{Trinh, X.S.}, \bibinfo{author}{Ngoduy, D.}, \bibinfo{author}{Keyvan-Ekbatani, M.}, \bibinfo{author}{Robertson, B.}, \bibinfo{year}{2022}.
\newblock \bibinfo{title}{Incremental unscented kalman filter for real-time traffic estimation on motorways using multi-source data}.
\newblock \bibinfo{journal}{Transportmetrica A: Transport Science} \bibinfo{volume}{18}, \bibinfo{pages}{1127--1153}.
\bibitem[{Tu et~al.(2021)Tu, Xiao, Li and Fu}]{tu2021estimating}
\bibinfo{author}{Tu, W.}, \bibinfo{author}{Xiao, F.}, \bibinfo{author}{Li, L.}, \bibinfo{author}{Fu, L.}, \bibinfo{year}{2021}.
\newblock \bibinfo{title}{Estimating traffic flow states with smart phone sensor data}.
\newblock \bibinfo{journal}{Transportation research part C: emerging technologies} \bibinfo{volume}{126}, \bibinfo{pages}{103062}.
\bibitem[{Underwood(1960)}]{underwood1960speed}
\bibinfo{author}{Underwood, R.T.}, \bibinfo{year}{1960}.
\newblock \bibinfo{title}{Speed, volume, and density relationships} .
\bibitem[{Wang et~al.(2016)Wang, Fan and Work}]{wang2016efficient}
\bibinfo{author}{Wang, R.}, \bibinfo{author}{Fan, S.}, \bibinfo{author}{Work, D.B.}, \bibinfo{year}{2016}.
\newblock \bibinfo{title}{Efficient multiple model particle filtering for joint traffic state estimation and incident detection}.
\newblock \bibinfo{journal}{Transportation Research Part C: Emerging Technologies} \bibinfo{volume}{71}, \bibinfo{pages}{521--537}.
\bibitem[{Wang et~al.(2021)Wang, Chen and Qu}]{wang2021model}
\bibinfo{author}{Wang, S.}, \bibinfo{author}{Chen, X.}, \bibinfo{author}{Qu, X.}, \bibinfo{year}{2021}.
\newblock \bibinfo{title}{Model on empirically calibrating stochastic traffic flow fundamental diagram}.
\newblock \bibinfo{journal}{Communications in transportation research} \bibinfo{volume}{1}, \bibinfo{pages}{100015}.
\bibitem[{Wang et~al.(2023)Wang, Wu, Zhuang and Sun}]{wang2023low}
\bibinfo{author}{Wang, X.}, \bibinfo{author}{Wu, Y.}, \bibinfo{author}{Zhuang, D.}, \bibinfo{author}{Sun, L.}, \bibinfo{year}{2023}.
\newblock \bibinfo{title}{Low-rank hankel tensor completion for traffic speed estimation}.
\newblock \bibinfo{journal}{IEEE Transactions on Intelligent Transportation Systems} \bibinfo{volume}{24}, \bibinfo{pages}{4862--4871}.
\bibitem[{Wang and Papageorgiou(2005)}]{wang2005real}
\bibinfo{author}{Wang, Y.}, \bibinfo{author}{Papageorgiou, M.}, \bibinfo{year}{2005}.
\newblock \bibinfo{title}{Real-time freeway traffic state estimation based on extended kalman filter: a general approach}.
\newblock \bibinfo{journal}{Transportation Research Part B: Methodological} \bibinfo{volume}{39}, \bibinfo{pages}{141--167}.
\bibitem[{Wang et~al.(2022)Wang, Zhao, Yu, Hu, Zheng, Hua, Zhang, Hu and Guo}]{wang2022real}
\bibinfo{author}{Wang, Y.}, \bibinfo{author}{Zhao, M.}, \bibinfo{author}{Yu, X.}, \bibinfo{author}{Hu, Y.}, \bibinfo{author}{Zheng, P.}, \bibinfo{author}{Hua, W.}, \bibinfo{author}{Zhang, L.}, \bibinfo{author}{Hu, S.}, \bibinfo{author}{Guo, J.}, \bibinfo{year}{2022}.
\newblock \bibinfo{title}{Real-time joint traffic state and model parameter estimation on freeways with fixed sensors and connected vehicles: State-of-the-art overview, methods, and case studies}.
\newblock \bibinfo{journal}{Transportation Research Part C: Emerging Technologies} \bibinfo{volume}{134}, \bibinfo{pages}{103444}.
\bibitem[{Wang et~al.(2024)Wang, Wang, Jia, Zhang, Klimenko, Wang, He, Huang and Liu}]{wang2024spatiotemporal}
\bibinfo{author}{Wang, Z.}, \bibinfo{author}{Wang, Y.}, \bibinfo{author}{Jia, F.}, \bibinfo{author}{Zhang, F.}, \bibinfo{author}{Klimenko, N.}, \bibinfo{author}{Wang, L.}, \bibinfo{author}{He, Z.}, \bibinfo{author}{Huang, Z.}, \bibinfo{author}{Liu, Y.}, \bibinfo{year}{2024}.
\newblock \bibinfo{title}{Spatiotemporal fusion transformer for large-scale traffic forecasting}.
\newblock \bibinfo{journal}{Information Fusion} \bibinfo{volume}{107}, \bibinfo{pages}{102293}.
\bibitem[{Wu et~al.(2024)Wu, Cheng, Chen, Qiu and Sun}]{wu2024traffic}
\bibinfo{author}{Wu, F.}, \bibinfo{author}{Cheng, Z.}, \bibinfo{author}{Chen, H.}, \bibinfo{author}{Qiu, Z.}, \bibinfo{author}{Sun, L.}, \bibinfo{year}{2024}.
\newblock \bibinfo{title}{Traffic state estimation from vehicle trajectories with anisotropic gaussian processes}.
\newblock \bibinfo{journal}{Transportation Research Part C: Emerging Technologies} \bibinfo{volume}{163}, \bibinfo{pages}{104646}.
\bibitem[{Xu et~al.(2020)Xu, Wei, Peng, Xuan and Guo}]{xu2020ge}
\bibinfo{author}{Xu, D.}, \bibinfo{author}{Wei, C.}, \bibinfo{author}{Peng, P.}, \bibinfo{author}{Xuan, Q.}, \bibinfo{author}{Guo, H.}, \bibinfo{year}{2020}.
\newblock \bibinfo{title}{Ge-gan: A novel deep learning framework for road traffic state estimation}.
\newblock \bibinfo{journal}{Transportation Research Part C: Emerging Technologies} \bibinfo{volume}{117}, \bibinfo{pages}{102635}.
\bibitem[{Xue et~al.(2024)Xue, Ka, Feng and Ukkusuri}]{xue2024network}
\bibinfo{author}{Xue, J.}, \bibinfo{author}{Ka, E.}, \bibinfo{author}{Feng, Y.}, \bibinfo{author}{Ukkusuri, S.V.}, \bibinfo{year}{2024}.
\newblock \bibinfo{title}{Network macroscopic fundamental diagram-informed graph learning for traffic state imputation}.
\newblock \bibinfo{journal}{Transportation Research Part B: Methodological} , \bibinfo{pages}{102996}.
\bibitem[{Zhang et~al.(2024)Zhang, Mao, Yang, Ma, Li and Gao}]{zhang2024physics}
\bibinfo{author}{Zhang, J.}, \bibinfo{author}{Mao, S.}, \bibinfo{author}{Yang, L.}, \bibinfo{author}{Ma, W.}, \bibinfo{author}{Li, S.}, \bibinfo{author}{Gao, Z.}, \bibinfo{year}{2024}.
\newblock \bibinfo{title}{Physics-informed deep learning for traffic state estimation based on the traffic flow model and computational graph method}.
\newblock \bibinfo{journal}{Information Fusion} \bibinfo{volume}{101}, \bibinfo{pages}{101971}.
\bibitem[{Zhao and Yu(2023)}]{zhao2023observer}
\bibinfo{author}{Zhao, C.}, \bibinfo{author}{Yu, H.}, \bibinfo{year}{2023}.
\newblock \bibinfo{title}{Observer-informed deep learning for traffic state estimation with boundary sensing}.
\newblock \bibinfo{journal}{IEEE Transactions on Intelligent Transportation Systems} .
\bibitem[{Zheng et~al.(2018)Zheng, Jabari, Liu and Lin}]{zheng2018traffic}
\bibinfo{author}{Zheng, F.}, \bibinfo{author}{Jabari, S.E.}, \bibinfo{author}{Liu, H.X.}, \bibinfo{author}{Lin, D.}, \bibinfo{year}{2018}.
\newblock \bibinfo{title}{Traffic state estimation using stochastic lagrangian dynamics}.
\newblock \bibinfo{journal}{Transportation Research Part B: Methodological} \bibinfo{volume}{115}, \bibinfo{pages}{143--165}.
\bibitem[{Zou et~al.(2023)Zou, Lai, Ma, Li and Wang}]{zou2023novel}
\bibinfo{author}{Zou, G.}, \bibinfo{author}{Lai, Z.}, \bibinfo{author}{Ma, C.}, \bibinfo{author}{Li, Y.}, \bibinfo{author}{Wang, T.}, \bibinfo{year}{2023}.
\newblock \bibinfo{title}{A novel spatio-temporal generative inference network for predicting the long-term highway traffic speed}.
\newblock \bibinfo{journal}{Transportation research part C: emerging technologies} \bibinfo{volume}{154}, \bibinfo{pages}{104263}.

\end{thebibliography}

\end{sloppypar}
\end{document}